\documentclass[sigconf]{acmart}

\AtBeginDocument{%
  \providecommand\BibTeX{{%
    \normalfont B\kern-0.5em{\scshape i\kern-0.25em b}\kern-0.8em\TeX}}}

\usepackage{textcomp}
\usepackage{graphicx}
\usepackage{amsmath}
\usepackage{graphicx}
\usepackage{xspace}
\usepackage{multirow}
\usepackage{multicol}
\usepackage{float}
\usepackage{subcaption}

\usepackage{algorithm}
\usepackage{url}
\usepackage{algpseudocode}
\usepackage{hyperref}
\hypersetup{colorlinks=true,urlcolor=blue}

\allowdisplaybreaks

\newcommand{\figref}[1]{Fig.~\ref{#1}}

\newcommand{\tabref}[1]{Table~\ref{#1}}

\newcommand{\secref}[1]{Sec.~\ref{#1}}

\newcommand{\etal}{\emph{et~al.}\xspace}
\newcommand{\eg}{\emph{e.g.},\xspace}
\newcommand{\ie}{\emph{i.e.},\xspace}
\newcommand{\etc}{etc.\xspace}

\newcommand\fakeparagraph[1]{\par\noindent\textbf{{#1}}.\xspace}

\newcommand{\preLD}{\emph{pre-LD}\xspace}
\newcommand{\LD}{\emph{LD}\xspace}

\usepackage[colorinlistoftodos,textsize=tiny,textwidth=15mm]{todonotes}

\renewcommand\footnotetextcopyrightpermission[1]{} 
\begin{document}

\title{Technical Report: Transferable Models to Understand the\\ Impact of Lockdown Measures on Local Air Quality}

\author{Johanna Einsiedler$^{1}$, Yun Cheng$^3$, Franz Papst$^{2}$,  Olga Saukh$^{2}$}
\affiliation{%
  	\institution{$^{1}$Vienna University of Business Economics / CSH Vienna, Austria}
	\institution{$^{2}$Graz University of Technology / CSH Vienna, Austria}
    	\institution{$^{3}$Computer Engineering and Networks Lab, ETH Zurich, Switzerland}
	\institution{\texttt{johanna\_einsiedler@gmx.at, chengyu@ethz.ch, \{papst, saukh\}@tugraz.at}}
}

\begin{abstract}
The COVID-19 related lockdown measures offer a unique opportunity to understand how changes in economic activity and traffic affect ambient air quality and how much pollution reduction potential can the society offer through digitalization and mobility-limiting policies. In this work, we estimate pollution reduction over the lockdown period by using the measurements from ground air pollution monitoring stations, training a long-term prediction model and comparing its predictions to measured values over the lockdown month. We show that our models achieve state-of-the-art performance on the data from air pollution measurement stations in Switzerland, Austria and in China: evaluate up to -29.4\% / +9.8\% and -22.6\% / +11.5\%, change in NO2 / PM10 in Eastern Switzerland and Lower Austria respectively;   -28.1\%, / -10.8\,\% and -52.8\,\% / -50.0\,\% in NO2 / PM2.5 in Beijing and Wuhan respectively. Our reduction estimates are consistent with recent publications, yet in contrast to prior works, our method takes local weather into account. What can we learn from pollution emissions during lockdown? The lockdown period was too short to train meaningful models from scratch. To tackle this problem, we use transfer learning to newly fit only traffic-dependent variables. We show that the resulting models are accurate, suitable for an analysis of the post-lockdown period and capable of estimating the future air pollution reduction potential.
\end{abstract}





\maketitle

\section{Introduction}
\label{sec:intro}



%

\begin{figure*}[t]
	\centering
	\subfloat[\footnotesize{NO2 changes pre/during/post lockdown period.}]{
		\includegraphics[width=0.5\textwidth]{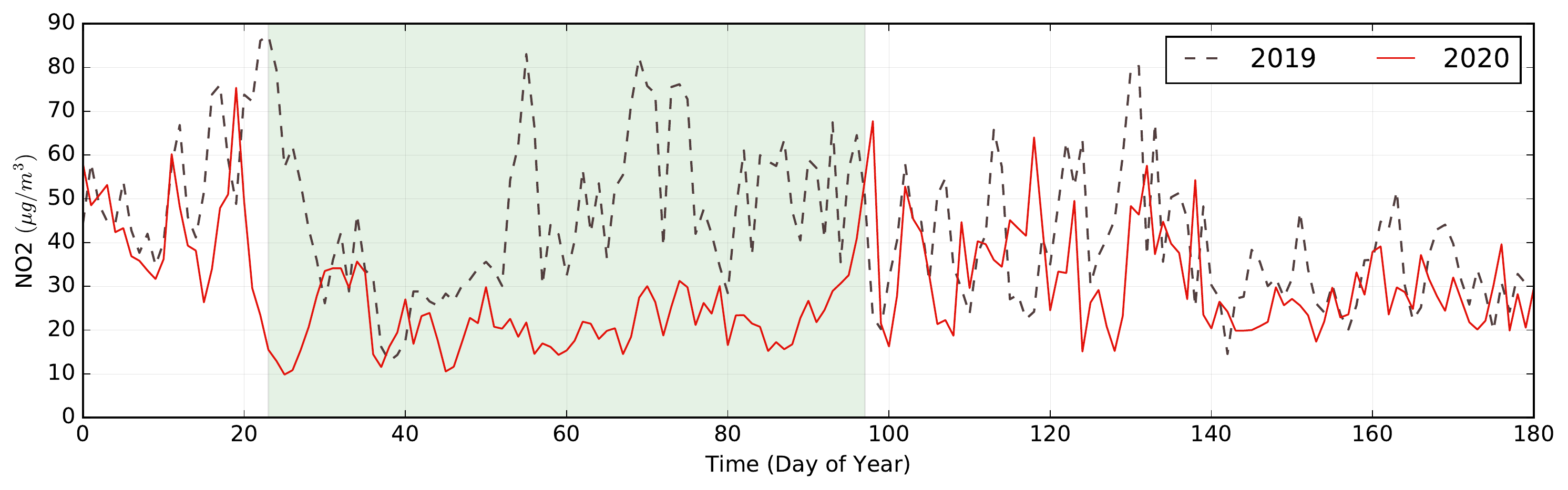}}
	\subfloat[\footnotesize{PM2.5 changes pre/during/post lockdown period.}]{
		\includegraphics[width=0.5\textwidth]{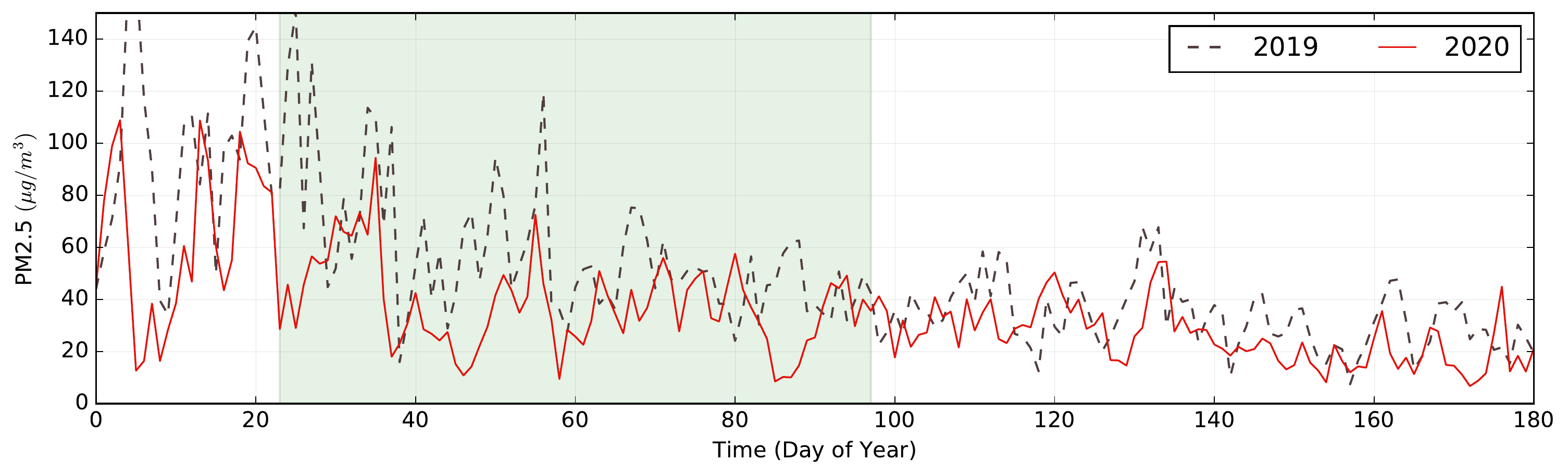}}
	\caption{Air pollution measured in 2019 and 2020 from Jan 1 to June 30 in Wuhan, China. Green zones show the lockdown period from Jan 23 to Apr 27, 2020. }
	\label{fig:intro}
\end{figure*}

Air quality  is of vital importance to human health. Medical studies have shown that PM2.5 (particulate matter of diameter less than 2.5 micron) can be easily absorbed by the lungs, and prolonged exposure may lead to respiratory impairments, blood diseases and neurodevelopmental disorders, such as autism, attention deficit disorders and cognitive delays~\cite{Cory19}. Air pollution is also found to have a negative effect on the cognitive functions in elderly adults~\cite{Ranft09} and is associated with increased mortality rates~\cite{Burnett18,Lelieveld20}. Furthermore, air pollution leads to enormous economic losses~\cite{OECD19} and its reduction is particularly important in overpopulated urban areas.

In the context of current pandemic, recent studies show that long-term exposure to PM2.5 and NO2 increases human susceptibility to SARS-CoV-2~\cite{Han20,Lelieveld20} and contributes to higher fatality rates~\cite{Ogen20, Wu20}. There is also a worrying evidence that the virus can be found in outdoor particulate matter~\cite{Setti20}. Although further investigations are necessary, it seems obvious that air pollution plays an important role with regard to both transmission and severity of this disease. Lockdown measures of varying duration and strictness in response to COVID-19 have shown to be effective to slow down the virus spread in winter and spring 2020 in many countries. At the same time, reduced mobility, working from home, accelerated digitalization and e-commerce made researchers wonder about the pollution reduction potential also in the context of global warming and while preserving the basic operations of cities and counties.

The 2020 lockdown provides a unique and valuable opportunity to analyze the air quality reduction patterns. \figref{fig:intro} presents a comparison of measured air pollution in Wuhan, China over the same period of time in 2019 and 2020.
Due to significantly reduced human activities, such as traffic, the concentrations of NO2 drop to low and stable levels during the lockdown compared to the same period in 2019 as shown in \figref{fig:intro}. Similar patterns are also observed for other pollutants, \eg for PM2.5 in Wuhan, China.

Although numerous studies estimate pollution reduction during lockdown in various countries~\cite{Grange20,Venter20,CREA20,Han20}, the results mostly represent aggregated differences to various baselines with no modelling of the dependency between air pollution and the local weather conditions, time-varying and land-use information. Moreover, a detailed analysis of the lockdown is difficult due to its short duration and, thus, scarceness of the representative data. This paper tackles the problem by making use of the air quality and weather data measured by public stations operated by official authorities. We also propose a modelling framework that enables such analysis, and give arguments for its usefulness in broader contexts.

Today, air pollution is measured by networks of governmental stations packed with expensive analytical instruments, satellite images~\cite{Bauwens20}, IoT devices equipped with low-cost sensors~\cite{saukh2013rendezvous}, passive samplers, crowdsourcing campaigns~\cite{hasenfratz2012participatory}, \etc. Real-time data from public and private stations can often be found online, \eg~\cite{Standort}. Numerous models were developed ranging from country-scale~\cite{ricke2018country} to city-scale~\cite{Hasenfratz15,mueller2016stat, hasenfratz2014percom} and aiming to predict short-term~\cite{zhang2019multi} and long-term pollution exposure~\cite{Barmpadimos11}. Although there is a large body of literature investigating the relationship between industry, traffic and air pollution, it is still not well understood how \emph{changes} in economic activity and traffic affect ambient air quality~\cite{wang2019effect}.

\fakeparagraph{Challenges}
The COVID-19 related lockdown measures offer a unique chance to build more knowledge in this area. However, there are numerous challenges to be solved. (1) Recent studies investigating the impact of the lockdown measures on air pollution are not correcting for the influence of weather conditions on air pollution, which can considerably distort the obtained estimates. (2) Strict initial lockdown measures took place only for a few weeks in countries around the world, which complicates learning a reasonable model for the lockdown period. Later lockdowns incorporated a different set of measures and thus can not be used to enhance the training data. Solving the first challenge helps to accurately compute the local pollution reductions over the lockdown period and understand their spatio-temporal variability. Solving the second challenge enables learning from the lockdown experience by computing different scenarios, such as estimating the air pollution reduction due to a partial back-to-normal regime or predicting pollution patterns if the lockdown would have happened during a different season or if its duration would have been extended.

\fakeparagraph{Contributions and road-map}
IIn this paper, we solve the above challenges by building \emph{the first long-term predictive models for the lockdown period} (\LD models). We use the following pipeline to achieve the goal: Based on the historical data over the past several years before the lockdown, we train long-term interpretable pre-lockdown (\preLD) models based on the Generalized Additive Models (GAMs) and show that they achieve comparable accuracy to the long-term and short-term models described in the literature and evaluated on the same data. The \preLD models are used to predict air pollution for the lockdown period while taking weather conditions into account. The predictions are then compared to the actually measured values over the same period. In addition, we leverage the additive property of the GAMs and their interpretability to train weather-aware \LD models using the scarce lockdown data. Towards this end, we fix environmental dependencies in the models and use transfer learning to compute a new fit for only land-use and daytime dependent parameters. We show that scarce data over only 4 weeks of lockdown are sufficient to train high-quality \LD models for NO2. We use both model classes to analyze the post-lockdown data and to estimate air pollution reduction in different scenarios. Our approach is evaluated on two public data sets from China and Switzerland. The code of the models is publicly shared on GitHub\footnote{https://github.com/johanna-einsiedler/covid-19-air-pollution}. More generally, we argue that a combination of the model interpretability and transfer learning notably simplifies result validation and increases their trustworthiness.
\secref{sec:related} summarizes a rapidly growing body of related works on modelling air pollution exposure.


\section{Related Work}
\label{sec:related}

Big data has a huge impact on modelling environmental processes~\cite{Lyon2015,Pietsch2017}, including air pollution. Even though the Earth generates data at a fixed pace, which doesn't change no matter how much data is collected~\cite{rolnick2019}, the already collected large volumes of observational data have been successfully used for modelling environmental processes~\cite{Faghmous2014,Caldwell2014,Sprenger2017}. Long-term environmental predictions are, however, largely rooted in scientific theory, which is one of the key reasons for their predictive power~\cite{rolnick2019}. Below we summarize the related works on theory-free and theory-based data-driven models when modelling air pollution and discuss the challenges we face when applying these to modelling air pollution under the COVID-19 lockdown measures.

\fakeparagraph{Interpretable air pollution models}
Classical dispersion models~\cite{zhang2012real} are still widely used for air quality mid-term and long-term predictions and interpretable analysis.
These models identify the root cause of air pollution from chemical, emission, climatological factors and combinations thereof. These models are described by a numerical function of emissions from the industry, traffic, meteorology, and other factors. A fitted model can then be used to understand the impact of each of these factors in isolation.

Among the model-based predictive environmental models, GAMs have shown to be able to facilitate a high degree of accuracy while retaining explainability. Thus GAMs have been frequently used to model air pollution~\cite{Hasenfratz15,Bertaccini12,Pearce11,Belusic15} and run interpretability analysis. For example, estimating the impact of traffic and weather on PM and NO2~\cite{Bertaccini12} and quantifying the impact of weather on NO2, PM and O3 for Melbourne~\cite{Pearce11}. Belusic~\etal~\cite{Belusic15} further analyze the impact of meteorological variables numerically in a model and explain 45\% of variance in CO, 14\% in SO2, 25\% in NO2 and 24\% in PM10. 
We apply the model selection procedure for GAMs~\cite{Barmpadimos11} to find the best model hyper-parameters and use log-normal pollutant values as an independent variable~\cite{Limpert01} for better prediction quality. Moreover, we leverage the additive property of GAM models to tackle data scarcity when training the \LD model.



\fakeparagraph{Short-term air pollution exposure prediction}
Recent research results on short-term air quality prediction range from a few hours to a few days ahead and mainly rely on deep learning models. FFA~\cite{zheng2015forecasting} is one of the first model-free data-driven methods which forecasts air quality from meteorological and weather inputs. DeepAir~\cite{yi2018deep} was proposed to learn the air pollution patterns in a deep manner, simultaneously considering individual and holistic influences. To further improve the model capacity, GeoMAN~\cite{liang2018geoman} used a three stage attention model learned from local features, global features and temporal geo-sensory time series. This approach shows a potential to learn the dynamic spatio-temporal correlations and to interpret the model results. Lin~\etal~\cite{lin2018exploiting} represent the spatial correlations in a graph with automatically selected important geographic features that affect PM2.5 concentrations, and use these features to compute the adjacency graph for the model. To conquer the challenge of sample scarcity, Chen~\etal~\cite{chen2019deep} proposed a multi-task approach to learn the representations from the relevant spatial and sequential data, as well as to build the correlation between air quality and these representations. Zhang~\etal~\cite{zhang2019multi} found that local fine-grained weather data is helpful to predict air quality. Their method fuses heterogeneous weather, air quality and Point-of-Interest (POI) data to learn the interactions between different feature groups. Ensemble methods, such as the winning solution of the air quality prediction challenge at KDD Cup 2018~\cite{luo2019accuair} are also used to further improve the accuracy of short-term air quality predictions. However, these deep learning approaches lack the ability to interpret the prediction results and usually focus on short-time horizons of a few hours to a few days.

\fakeparagraph{Transferable models}
Transfer learning~\cite{pan2009survey} promises to light-retrain a model in order to adapt the parameters to a changed setting and requires little data. Pollution models are usually very local and not spatially transferable, \eg across cities and countries, because the learned dependencies are location-specific and policies may vary substantially across distant areas. To transfer the knowledge from a city with sufficient multi-modal data and labels to a new city with data scarcity, Wei~\etal~\cite{wei2016transfer} propose a method to learn semantically related dictionaries from a source domain, and simultaneously transfer these dictionaries and labelled instances to the target domain. Temporal transferability of learned dependencies is difficult due to environment and policy changes over time. For example, Cheng~\etal~\cite{cheng2020maptransfer} applies a learning-based approach to solve the downscaled sensor deployment problem. They try to transfer knowledge from a historical dense deployment to current sparse deployment setting by finding the most similar instances to execute the model transfer. Transferring pollution models in space or in time requires strong assumptions about the source and the target domains, such as shared similarities and other transferable structures. Without explicit assumptions or known structural similarities, a model should be trained from scratch, which is data-intensive.

\fakeparagraph{Impact of COVID-19 on air quality}
Lockdown measures in response to COVID-19 pandemic offer a unique opportunity to improve prediction of policy impacts reinforcing work-from-home and changing to low-emission mobility vehicles such as bicycles. The study in \cite{Muhammad20} assessed NO2 reduction based on satellite imagery by NASA and ESA in multiple COVID-19 epicenters. A similar assessment of the impact of SARS-CoV-2 in other areas is provided in \cite{Bauwens20}. The relationship between air pollution and lockdown measured was studied in \cite{Venter20} using satellite data and ground level sensor data. The weather adjustment is taken into account but modelled as a simple linear dependency. A recent report~\cite{Thieriot20} estimates NO2 reduction for major European cities during the lockdown when compared to previous years. The results suggest a reduction from 16-18\,\% in Budapest and Berlin to over 60\,\% in Paris and Bucharest.

Building a good predictive model for the lockdown period is challenging due to a short lockdown duration of only several weeks in most countries. In contrast to all previous efforts, we are one of a few to provide a weather and season-compensated estimate of pollution reduction over the lockdown period\footnote{The only similar study conducted in parallel to this work, using Random Forests modelling and reporting similar results for Switzerland can be found here: \url{https://empa-interim.github.io/empa.interim/swiss_air_quality_and_covid_19.html} visited 2020-10-26.}, and we are the first to use transfer learning to train an interpretable model for the COVID-19 lockdown period valuable, for the analysis of the future air pollution reductions due to policy change and technological progress.

\section{Overall Framework}
We first describe the data sets used in the paper and then present the overall modelling framework.

\begin{table}
\begin{center}
\scriptsize{
\begin{tabular}{l|c|c|cccccc}
\toprule
Country & Class & Local situation &  \# stations \\
\midrule
\multirow{3}{*}{East Switzerland} 
    & No Traffic & Located offside the road & 1 \\
    & Low Traffic &  $\leq$30,000 VPD & 3 \\
    & High Traffic & $>$30,000 VPD & 1 \\
\hline
\multirow{4}{*}{China}
    & Urban     & Urban Beijing, parks  & 12 \\
    & Rural     & Countryside, parks    & 11 \\
    & Suburban  & Polluted transfer zones  & 7 \\
    & Road     & Urban, high traffic   & 5 \\
\hline
\multirow{4}{*}{Lower Austria}
    & Urban     & City, industry \& office areas  & 4 \\
    & Rural     & Fields    & 9 \\
    & Rural Residential  & Residential Areas outside of cities  & 6 \\
    & Residential     & Residential Areas in cities   & 8  \\
\bottomrule
\end{tabular}
}
\caption{Classification of stations in Switzerland by \#vehicles per day (VPD) and in Austria \& China by location type for NO2.}
\label{tab:che_classification}
\end{center}
\end{table}

\section{Data Sets}
\label{sec:data}

This section describes the data sets we use to train and test the \preLD and \LD models for Eastern Switzerland, Lower Austria, Beijing and Wuhan. We also shortly describe the progress and the duration of the lockdown measures. Note that the investigated countries implemented very different air pollution reduction policies over the years. Also the severity of the lockdown measures varied considerably. Both facts highlight robustness of the modelling approach presented in the paper.

\fakeparagraph{Beijing and Wuhan}
We collect air quality data\footnote{\url{https://quotsoft.net/air/} visited 2020-10-12}, including PM2.5, PM10, O3, NO2, CO and SO2, from 35 stations in Beijing and 10 stations in Wuhan from Jan 1, 2016 to Feb 5, 2021. Our scripts also fetch meteorological data\footnote{\url{https://darksky.net/} visited 2020-10-12} for the same station locations every hour during the same period of time. Each record comprises the following parameters:  weather situation (sunny, cloudy, overcast, foggy, snow, small rain, moderate rain, and heavy rain), relative humidity, temperature, pressure, wind speed, and wind direction data. The initial lockdown periods in Beijing and in Wuhan were between Jan 23 and Apr 8, 2020. Further lockdowns were considerably lighter or affected only local areas.
According to the local environmental situation for each air quality station, the stations in Beijing are categorized into four classes as shown in \tabref{tab:che_classification}: \emph{Road}, \emph{Rural}, \emph{Suburban} and \emph{Urban}.

\fakeparagraph{Eastern Switzerland}
Air quality data for eastern Switzerland are published by Ostluft\footnote{\url{https://www.ostluft.ch/} visited 2020-10-12}, an organisation measuring air quality in Eastern Switzerland and the Principality of Lichtenstein. The stations measure main air pollutants along with meteorological data. The set of modalities may vary from one station to another, although NO2, PM10 and O3 are measured by all stations. Stations are spread across the area, although bigger cities including Zurich and St.\,Gallen have more than one station at representative locations. We use the data from 5 stations measuring NO2, PM10, O3, CO, relative humidity, temperature, pressure, wind speed, and wind direction on the hourly basis from Jan 1, 2016 to Feb 5, 2021. TThe initial lockdown in Switzerland took place between Mar 16 and Apr 27, 2020~\cite{Bundesrat20,SRF20} and was much stricter than later measures.
According to the traffic conditions, the stations are classified into three classes\footnote{\url{https://www.ostluft.ch/index.php?id=19} visited 2020-10-26}: \emph{No Traffic}, \emph{Low Traffic} and \emph{High Traffic}. The details are summarized in~\tabref{tab:che_classification}.

\fakeparagraph{Lower Austria}
Data for Lower Austria was obtained from 'NUMBIS' \footnote{https://www.noe.gv.at/luft}. The dataset contained half-hourly observations from 1st of January 2016 to 31st of August 2020. It included 27 and 16 spatial measurement points for NO2 and PM10, respectively. The stations were grouped together according to the type of their location in four classes as shown in \tabref{tab:che_classification}: \emph{Urban}, \emph{Residential}, \emph{Rural Residential} and \emph{Rural}.
The first lockdown in Austria took place between March 20th and Apr 27th, 2020~\cite{Idowa20, NewsORF20}. 

In all considered areas the available data has hourly time resolution. However, we use daily aggregates to build and validate our models. We also note that the lockdown severity in Wuhan was the highest and in Switzerland the lowest.

\subsection{Framework}
\begin{figure}[t]
    \centering
    \includegraphics[width=.9\linewidth]{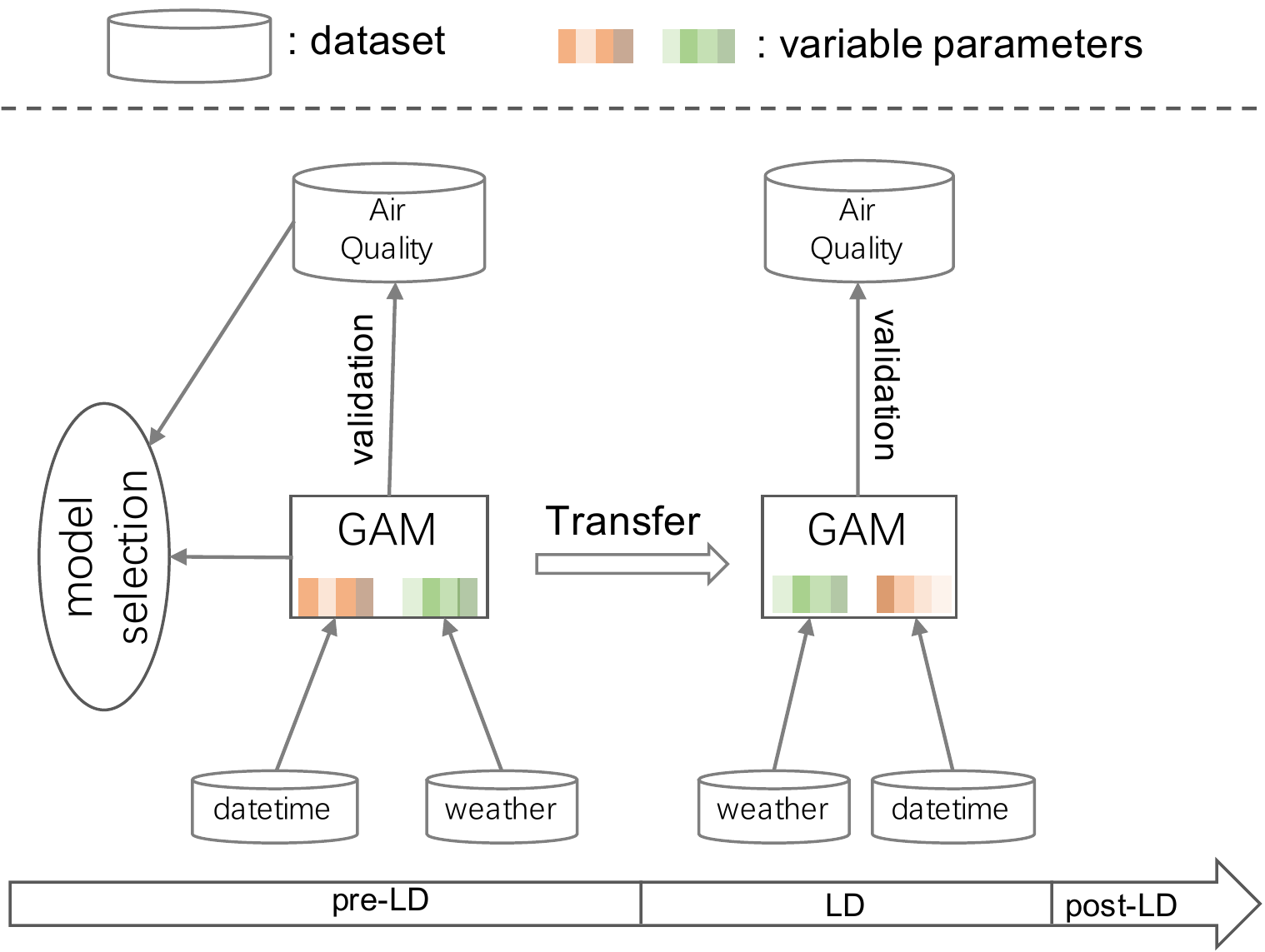}
    \caption{Overall framework.}
    \label{fig:framework}
\end{figure}

\figref{fig:framework} shows the overall framework of the proposed approach.
We rely on Generalized Additive Models (GAMs) due to their popularity, accurate prediction ability and model interpretability. These are described in \secref{sec:gam_intro}. We first use pre-lockdown (pre-LD) data comprising weather parameters and datetime features (see \secref{sec:gam_variables}) to train \preLD GAM models. As part of the training, we rely on the model selection algorithm described in \secref{sec:gam_selection} and validate the obtained models in \secref{sec:gam_validation}. The \preLD GAM models are then used to analyze the pollution reductions over the lockdown (LD) period. As next, due to data scarcity and the inherent inability to train GAM models from scratch for the LD period, we apply transfer learning mechanism to adapt the trained \preLD GAM models to the LD period (in \secref{sec:transferability}). We refer to the resulted models as \LD models. We leverage domain knowledge to justify the validity of the applied knowledge transfer: the impact of weather factors on air quality remains largely the same despite the lockdown. We therefore, fix the \preLD model parameters describing the weather dependency during the transfer procedure, and only retrain the datetime parameters that represent human mobility to obtain the \LD models. We show that the sparse data collected over the lockdown is sufficient to generate meaningful \LD models for NO2 using this approach.


\section{Interpretable Long-Term Air Pollution Predictive Models}
\label{sec:longtermmodel}

This section describes the process of training a \preLD model to be later used to estimate the air pollution level if no lockdown would have happened. We require a high degree of model interpretability for the following two reasons: (1) the prediction time horizon should cover the whole lockdown period of several weeks and thus the models should have sufficient predictive power for long-term predictions, and (2) model interpretability is essential to understand the impact of mobility drop on the reduction of air pollution and to learn from these conclusions. For these reasons, we adopt GAMs that have been successfully used to model air pollution in the past research~\cite{Hasenfratz15}, yet leverage additional optimizations~\cite{Barmpadimos11,Carslaw09} and statistical tests to ensure their robustness and optimize performance as described below. Since pollutant concentrations vary greatly depending on location and the surroundings of the monitoring station, a separate model was fitted for each station.

\subsection{Generalized Additive Models (GAMs)}
\label{sec:gam_intro}

GAMs have been proposed in 1986 by Hastie and Tibshirani~\cite{Hastie86} and blend properties of generalized linear models with additive models. The impact of the predictive variables is captured through non-parametric smooth functions. These are then summed up and related to the response variable via a link function:
\begin{equation}
g(E(Y)) = s_1(x_1) + s_2(x_2) + ... + s_p(x_p),
\end{equation}
\noindent where $E(Y)$ is the expected value of the dependent variable $Y$, $g(\cdot)$ is a link function between its argument and the expected value to the predictor variables $x_1, ..., x_p$, and $s_1(\cdot), ..., s_p(\cdot)$ denote non-parametric smooth functions. The statistical distribution of the concentration of air pollutants, similarly to many other environmental parameters, closely follows a log-normal distribution~\cite{Limpert01}. Thus, a logarithmic link function $g(\cdot)$ has been chosen similarly to \cite{Hasenfratz15}. The instance of our GAM model is therefore
\begin{eqnarray}
\ln(Y) &=&a + s_1(x_1) + s_2(x_2) + ... + s_p(x_p) + \\
       &&+ b_1\cdot Z_1 + b_2\cdot Z_2 + ... + b_q\cdot Z_q + \epsilon, \nonumber
\end{eqnarray}
\noindent where $a$ is the intercept, $Z_1, ..., Z_q$ denote categorical variables along with their respective weights $b_1, ..., b_q$ and $\epsilon$ is an error term.

\subsection{Explanatory Variables}
\label{sec:gam_variables}

The explanatory variables we use comprise meteorological variables: wind speed (WS), wind direction (WD), precipitation (P), temperature (T), dew point (DP) and relative humidity (RH). To ensure an accurate feature representation of the wind direction, the polar coordinates are transformed into cartesian coordinates:
\begin{equation}
\textrm{WD$_x$} = \sin \left( \frac{\textrm{WD}}{360} \cdot 2\pi \right),
\text{\hspace{.5cm}}
\textrm{WD$_y$} = \cos \left( \frac{\textrm{WD}}{360} \cdot 2\pi \right).
\end{equation}

\noindent Furthermore, an additional variable is created by applying principal component analysis (PCA) on precipitation, humidity, dew point and temperature. PCA is a dimensionality reduction method to reduce the mutual correlations between included variables. The first component is added to the set of explanatory variables as `PCA'.
We augment the set of explanatory variables with their lagged versions for one, two and three days. Wind speed (WS) and PCA are augmented with their respective rolling averages over the previous weeks. In addition, a categorical variable for month (M) and a variable for day of the year (DY)  is included to account for seasonal patterns. 

\subsection{Model Selection Algorithm}
\label{sec:gam_selection}
For the selection of the model covariates we use a forward elimination procedure. The algorithm closely follows the framework used in similar research designs in the environmental sciences~\cite{Barmpadimos11, Carslaw09}. Two key indicators are used for the model selection: the Akaike Information Criterion (AIC)~\cite{Akaike1974} and the Variance Inflation Factor (VIF)~\cite{Freund06}. The AIC is an estimate of the in-sample prediction error that is commonly used to compare the quality of different statistical models for a given data set~\cite{Hastie2001}. The aim of the indicator is to regularize the model by balancing the goodness-of-fit against model complexity and thereby avoiding both underfitting and overfitting. The AIC is calculated as follows:
\begin{equation}
AIC = 2k - \ln(l),
\end{equation}
\noindent where $k$ is the number of model parameters, and $l$ denotes the maximum value of the model likelihood.

The VIF measures the degree of collinearity between independent variables, \ie if they have a close to linear relationship and are thus not independent from each other. Collinearity may cause problems in regression-like techniques as it inflates the variance of regression parameters and thus may lead to wrong identification of the relevant predictors~\cite{Dormann12}. The VIF is calculated as follows:
\begin{equation}
VIF = 1 / (1-R^2_i),
\end{equation}
\noindent where $R^2_i$ is the coefficient of determination of the regression of the $i$-th variable with all other explanatory variables.

\fakeparagraph{Model selection algorithm}
Our implementation of the model selection algorithm closely follows \cite{Barmpadimos11}. The algorithm executes as follows: (1) For each explanatory variable we fit a GAM model comprising just this single variable. The model with the lowest AIC is selected. (2) We iteratively search for the next best variable to be added to the existing model. Variables with $VIF >2.5$ are filtered out. Among the constructed candidate models, the one with the lowest AIC is chosen. The threshold of $2.5$ corresponds to the coefficient of determination $R^2=0.6$. This conservative threshold setting was deemed appropriate taking into account the frequent interactions, and thus collinearity, between weather variables. Scientific papers dealing with weather data often adopt the threshold of 2.5~\cite{Barmpadimos11}, whereas higher cut-off values, \eg 4, 5 and 10 are found in the literature~\cite{Brien07, Carslaw09, Kutner05}. Step (2) is repeated until the addition of any other explanatory variable leads to an increase of AIC.

\begin{table*}
\begin{center}
\scriptsize{
\begin{tabular}{l|c|cccccccc}
\toprule
Variable & \multirow{2}{*}{Abbr.} & \multicolumn{2}{c}{Switzerland} &  \multicolumn{2}{c}{Austria} & \multicolumn{2}{c}{Beijing} & 
\multicolumn{2}{c}{Wuhan} \\
name && NO2 & PM10 & NO2 & PM2.5 & NO2 & PM2.5  \\ 
\midrule
    Wind speed          & WS        &4  &5 &24  &6 &32  &32 &5   &-  \\
    Wind direction X         & WD$_x$    &6 &8  &27 &24 &18  &21   &8   &10  \\
    Wind direction Y         & WD$_y$    &3  &4  &27 & 10 &7  &7&11   &11    \\
    Temperature           & T         &2  &5   &9 &14 &1   &-    &-   &- \\
    Relative humidity           & RH        &2   &3   &7 &10 &21 &9  &11   &9    \\
    Month               & M         &-  &-     &2 &- &-   &-   &-   &1 \\
    Day of the year          & DY   &2  &-    &13  &8 &24   &3   &4   &8 \\
    Dew point           & DP        &0  &- &1   &1 &7  &1 &1  &- \\
    PCA                 &PCA        &7  &-  &15 &2 &26 &21 &4  &9  \\
    Weekday             &D          &1  &2   &5  &- &-  &2  &1  &-\\
\bottomrule
\end{tabular}
\caption{\#stations where the corresponding explanatory variable was chosen by the model selection algorithm.}
\label{tab:model_selection_count}
}
\end{center}
\end{table*}

The results of the model selection algorithm for all stations in China and Switzerland are displayed in \tabref{tab:model_selection_count}. The value of each cell represents the frequency of the corresponding explanatory variable being selected into the GAM models. Only occurrences chosen by the model selection algorithm are listed in the table. Ultimately, the Weekday variable (D) was explicitly added to all models where it hasn't been automatically selected. This decision was made to specifically take into account weekly pollution periodicity patterns to reflect the traffic changes during the lockdown period. This technique has been used in a similar research design to analyse the long-term air pollution trends~\cite{Barmpadimos11}.

\subsection{Model Validation}
\label{sec:gam_validation}
\begin{figure*}[t]
	\centering
	\subfloat[\footnotesize{Switzerland NO2}]{
		\includegraphics[width=0.45\textwidth]{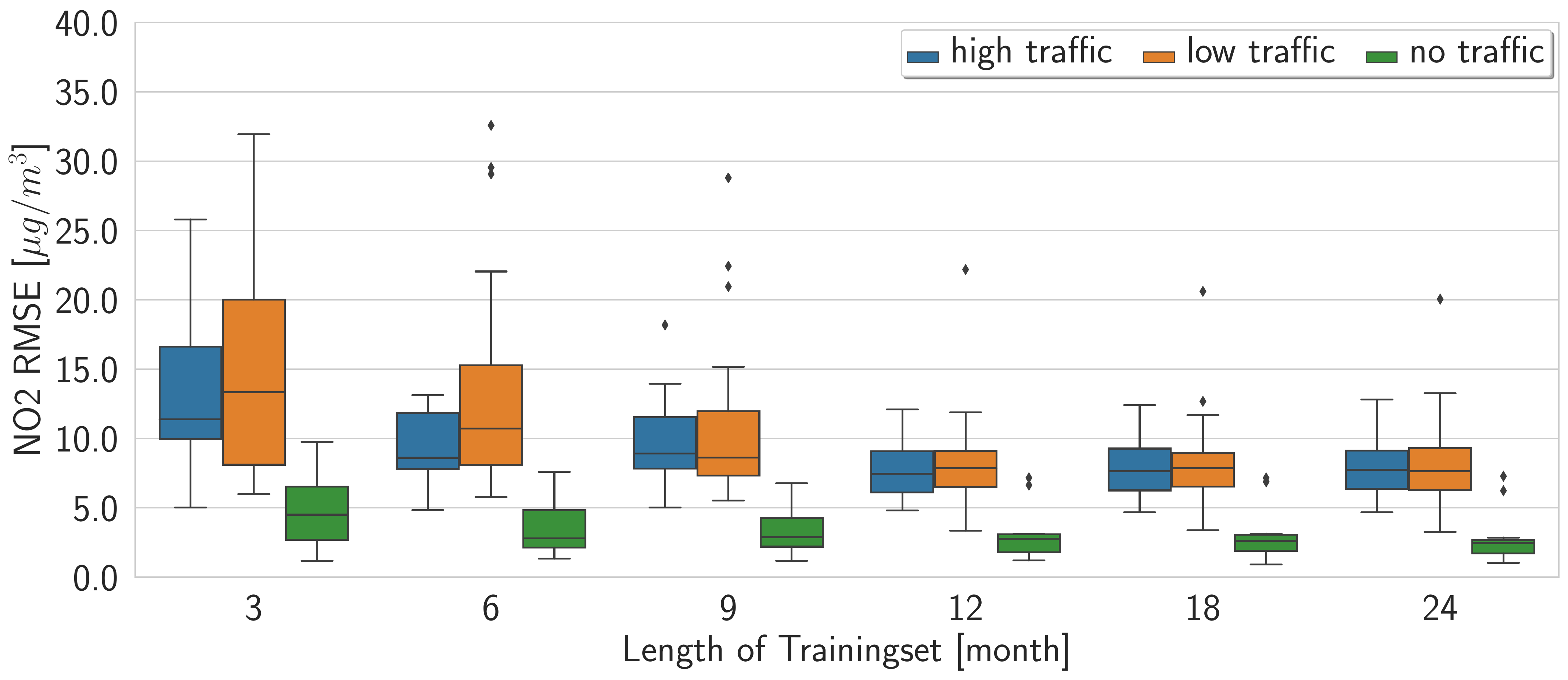}}
	\hspace{1cm}
	\subfloat[\footnotesize{Beijing NO2}]{
		\includegraphics[width=0.45\textwidth]{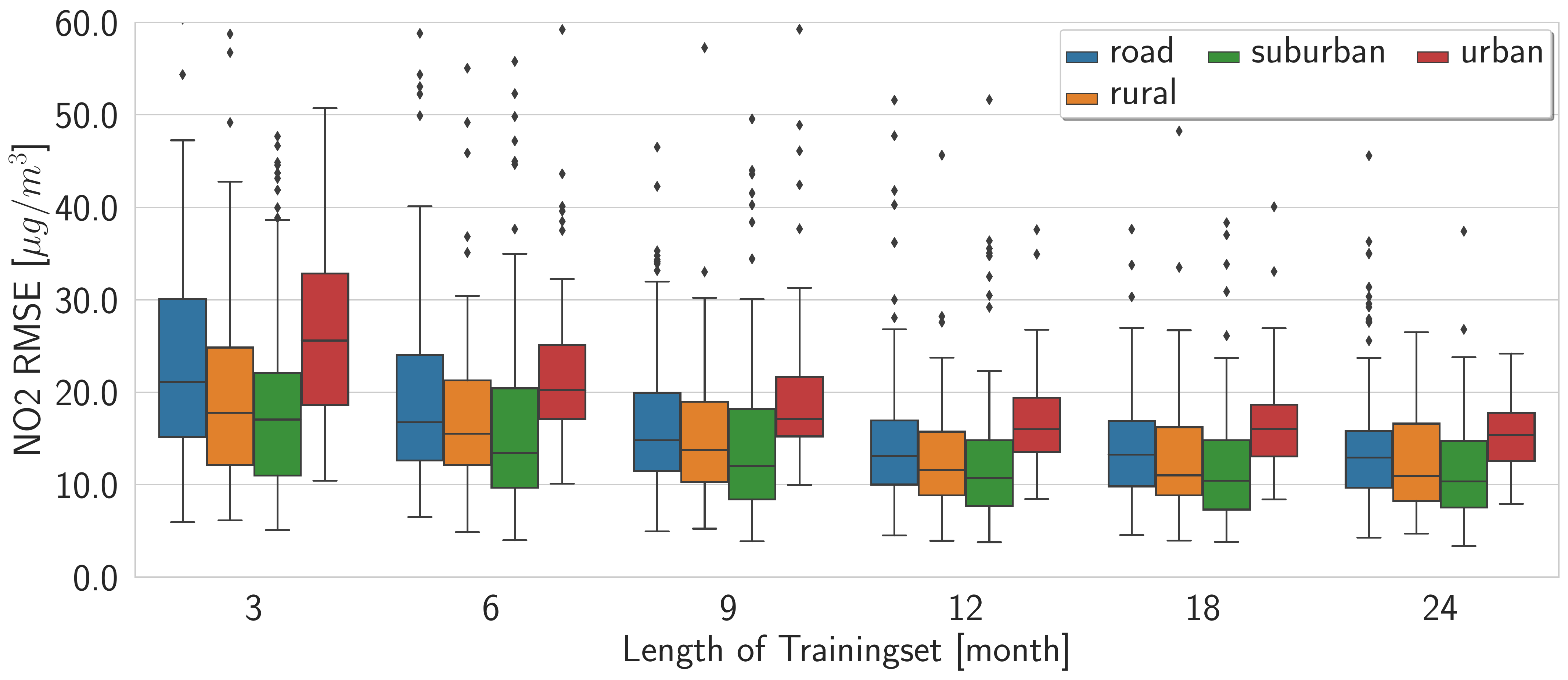}}
	
	\subfloat[\footnotesize{Switzerland PM10}]{
		\includegraphics[width=0.45\textwidth]{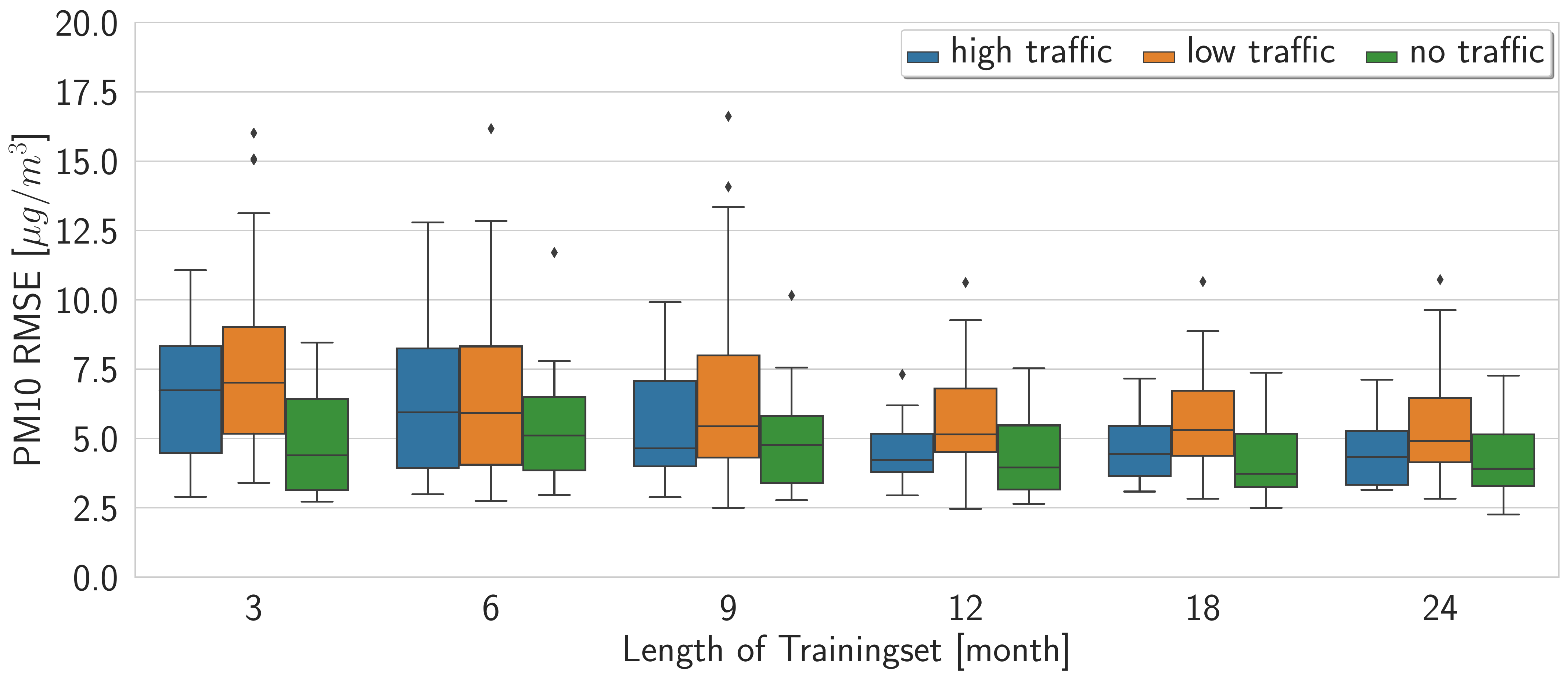}}
	\hspace{1cm}
	\subfloat[\footnotesize{Beijing PM2.5}]{
		\includegraphics[width=0.45\textwidth]{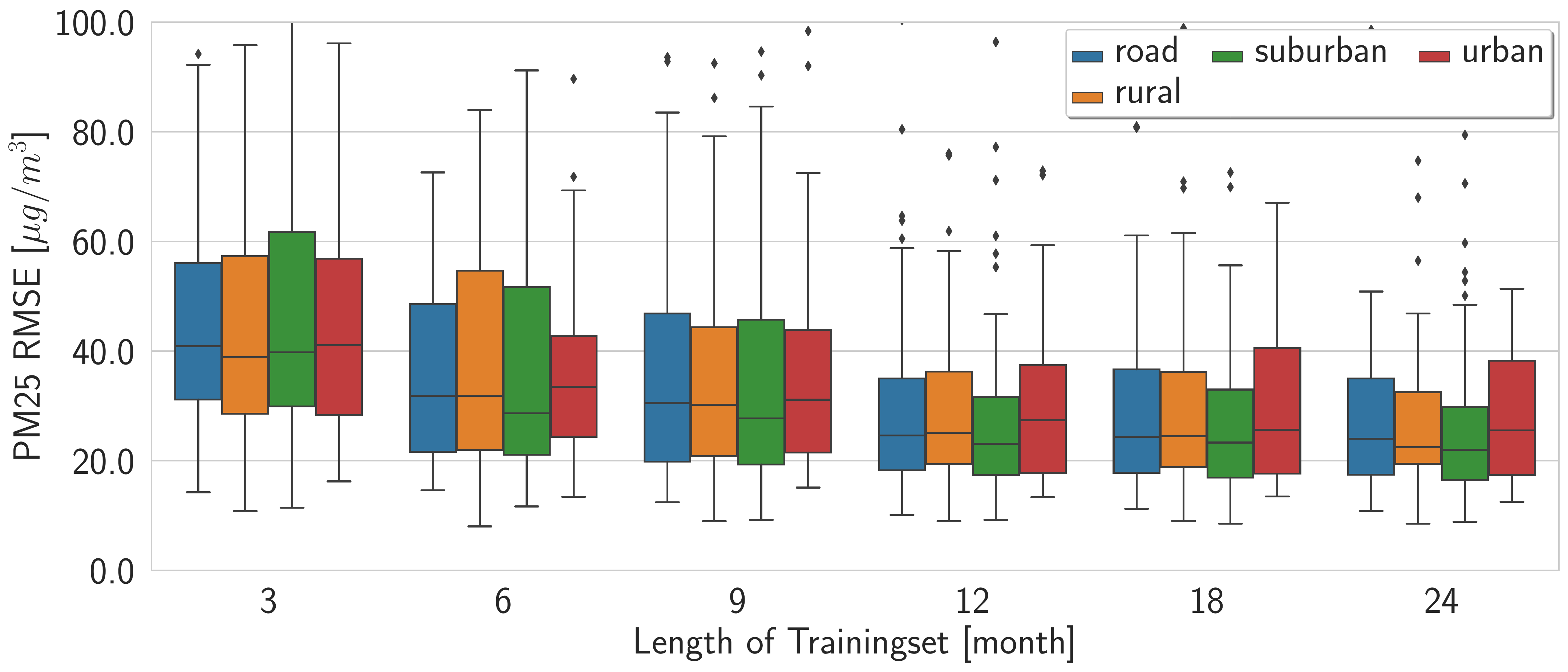}}
		
	\caption{\preLD model performance in cross-validation using different length of the train data.}
	\label{fig:validation}
\end{figure*}

We assess the quality of the trained \preLD models using cross-validation. The results for NO2 and PM for Switzerland, Beijing and Wuhan are shown in \figref{fig:validation}. Taking into account the temporal dependencies in the air pollution data, the models for different stations are fitted on 3, 6, 9, 12, 18 and 24 months of train data prior to a chosen date and tested on the data from the subsequent month. The chosen cut-off date is the start of each month in the year 2019. We observe that two years of data is necessary to train the \preLD GAM models of good quality. Considering more historical data does not substantially improve the model accuracy. For this reason, all further evaluations are based on the \preLD models trained on two years of data preceding the lockdown in each region.

\begin{table*}
\scriptsize{
\begin{center}
\subfloat[\preLD and \LD model performance in cross-validation.]{
    \begin{tabular}{l|l|cc|cc|cc|cc}
    \toprule
    \multirow{3}{*}{Model} & \multirow{3}{*}{Measure} & \multicolumn{2}{c}{Switzerland} &
    \multicolumn{2}{c}{Austria} &
    \multicolumn{2}{c}{Beijing} &
    \multicolumn{2}{c}{Wuhan} \\
    & & NO2 & PM10 & NO2 & PM10 & NO2 & PM2.5 & NO2 & PM2.5 \\
    & & $\mu g/ m^{3}$ & $\mu g/ m^{3}$ & $\mu g/ m^{3}$ & $\mu g/ m^{3}$ & $\mu g/ m^{3}$ & $\mu g/ m^{3}$ \\
    \midrule
    \multirow{2}{*}{\preLD}
    & RMSE          &7.16  &4.99 &4.5&8.12   &13.38   &29.41  & 14.61   &22.37 \\
    & R$^2$         &0.69  &0.54   &0.62& 0.55 &0.60   &0.55    &0.64   &0.66  \\
    \hline
    \LD 
    & RMSE          &7.03 &- & 3.68 &- &13.08 &-  &12.26 &-  \\
    \bottomrule
    \end{tabular}
    }
\subfloat[Comparison to related works.]{
    \begin{tabular}{l|c|cc}
    \toprule
    \multirow{4}{*}{Measure} &  Switzerland & \multicolumn{2}{c}{Beijing} \\
    & Barmpadimos~\etal~\cite{Barmpadimos11} & \multicolumn{2}{c}{Zhang~\etal~\cite{zhang2019multi}} \\
    & PM10 & PM2.5 [1-6h] & PM2.5 [19-24h] \\
    & $\mu g/ m^{3}$   & $\mu g/ m^{3}$ & $\mu g/ m^{3}$ \\
    \midrule
    RMSE    &2.60   &17.35 -- 25.81    & 26.88 -- 44.08 \\
    R$^2$   &0.62   &-    &-    \\
    \bottomrule
    \end{tabular}
}
\caption{\preLD and \LD model performance in cross-validation, comparison to related works.}
\label{tab:validation}
\end{center}
}
\end{table*}


\fakeparagraph{Eastern Switzerland}
For the Swiss stations the model has an average RMSE in cross validation of 7.16 for NO2 and 4.99 for PM10. Barmpadimos et al. \cite{Barmpadimos11} use GAM models fitted on detailed weather data to analyse PM10 trends in Switzerland. Their models fitted on 16 years of data reach a RMSE between 2.2 and 3.2 for PM10 for large test data sets, see \tabref{tab:validation}. We thus conclude that the quality of the obtained GAM models is comparable to  published results.

\fakeparagraph{Beijing}
For the stations in Beijing the average RMSE for PM2.5 in cross validation is 29.41. Zhang et. al. \cite{Zhang19} compare different models for \emph{short-term} (between 6\,h and 24\,h) PM2.5 predictions in Beijing over 2016-2018. The RMSE of these short-term models ranges between 26.9 and 44.1. Thus, our model matches well the performance of the current state-of-the-art predictive short-term models, while predicting a much longer time period and allowing for an immediate analysis of the fitted dependencies. In contrast to these short-term predictive models, however, our GAM models need much more (two years compared to a few days) training data to achieve acceptable accuracy. We note that historical weather data is often publicly available, which makes model training on large historical data sets possible.

In the next section, we use the trained \preLD models to estimate pollution reduction due to the COVID-19 lockdown measures.

\section{Impact of COVID19 on Air Pollution}
\label{sec:covid19}

\begin{figure*}[ht]
	\centering
	\subfloat[\footnotesize{Switzerland}]{
		\includegraphics[width=0.45\textwidth]{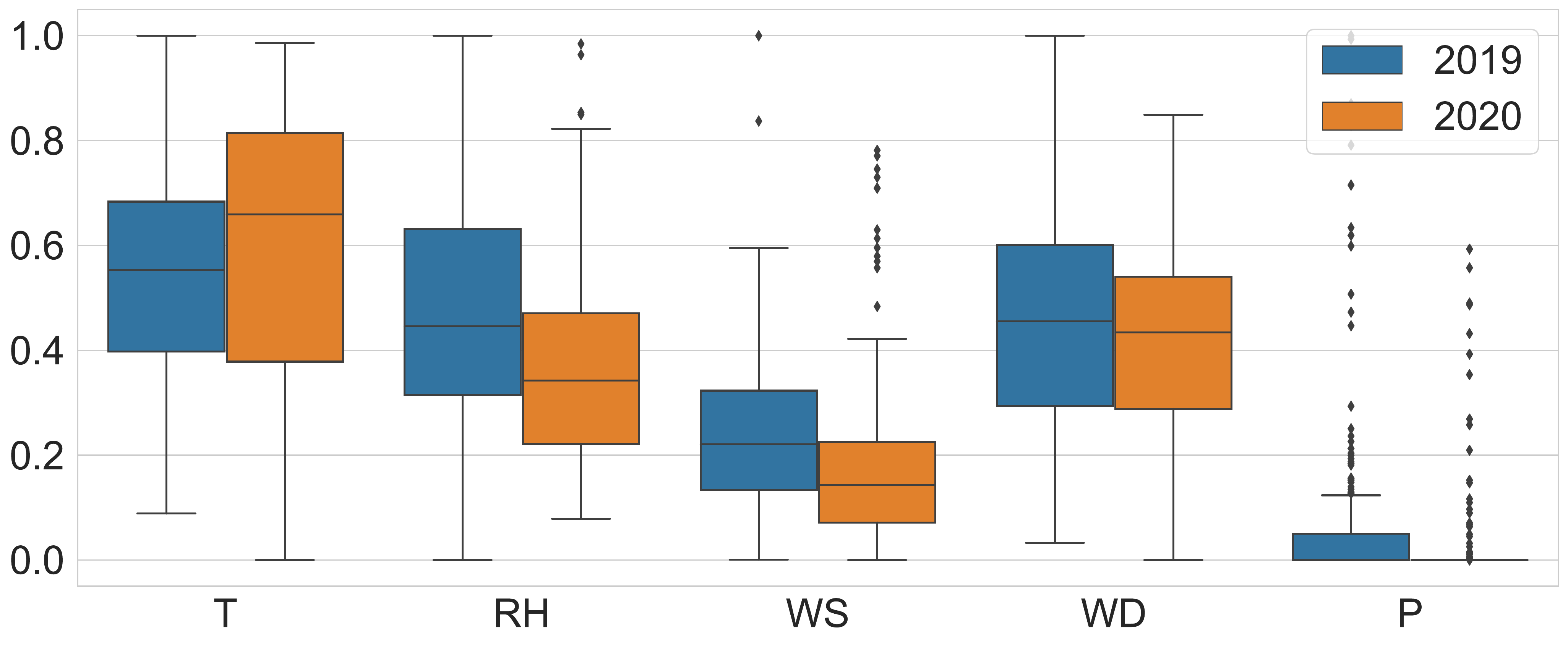}}
	\hspace{1cm}	
	\subfloat[\footnotesize{Austria}]{
		\includegraphics[width=0.45\textwidth]{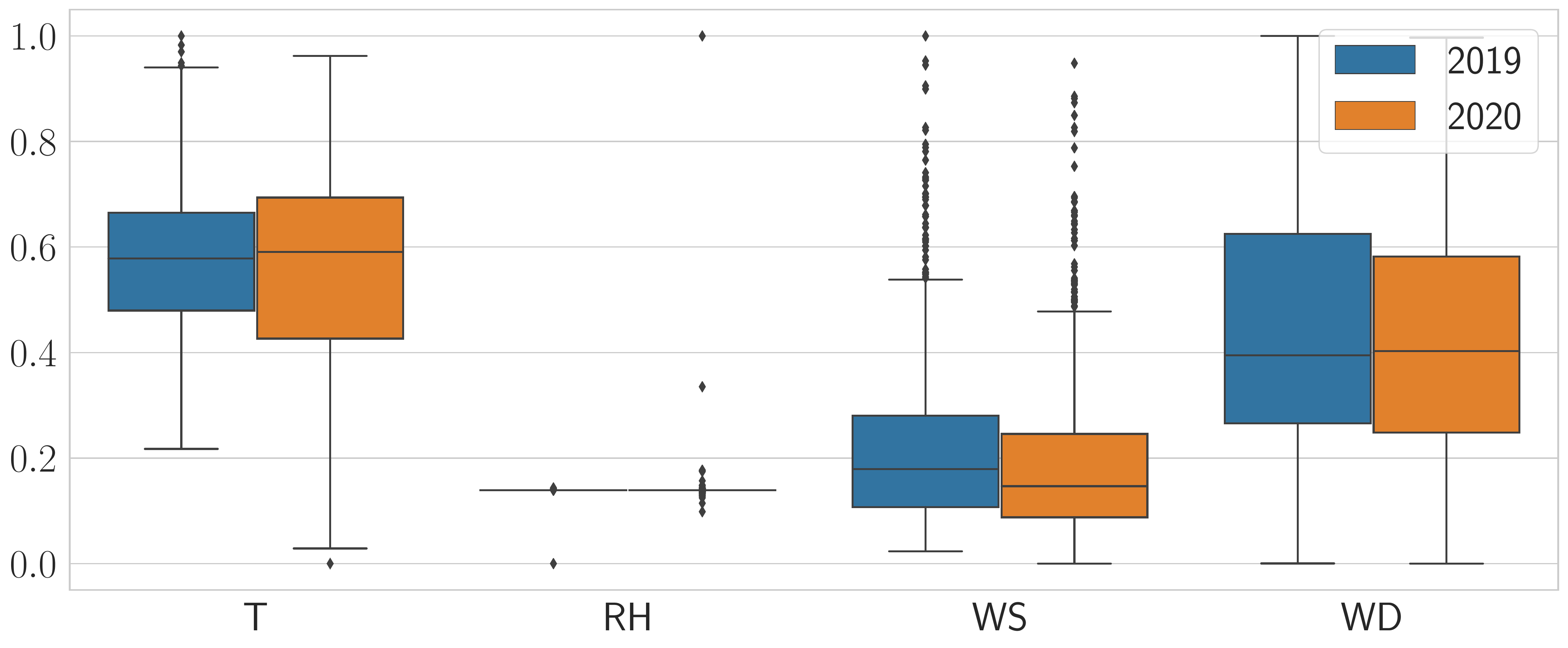}}
		
	\subfloat[\footnotesize{Beijing}]{
		\includegraphics[width=0.45\textwidth]{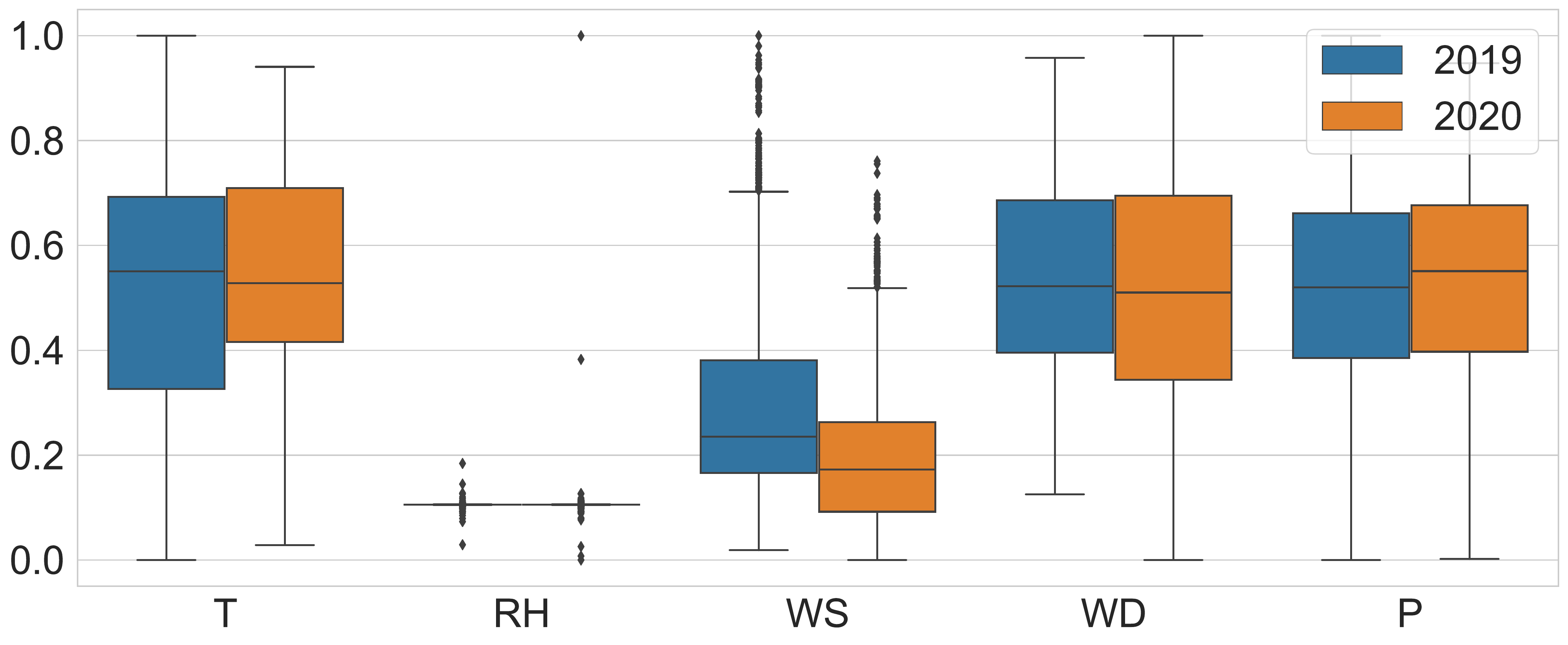}}
	\hspace{1cm}
	\subfloat[\footnotesize{Wuhan}]{
		\includegraphics[width=0.45\textwidth]{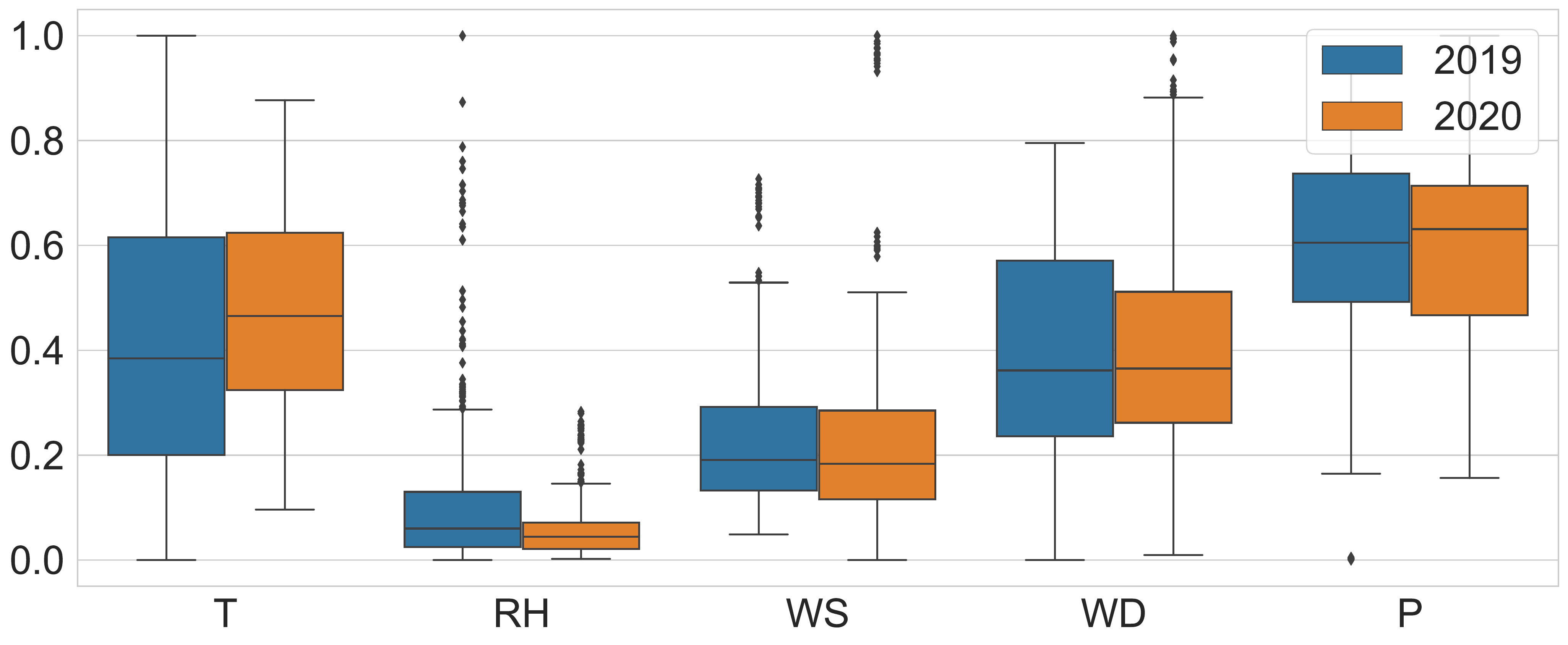}}
	\caption{Weather comparison for the lockdown periods in 2020 to the same period in 2019 in Switzerland, Beijing and Wuhan.}
	\label{fig:weather-comparison}
\end{figure*}

Conceptually, estimating the impact of intervention measures on air pollution shares similarities with  evaluating the impact of medical treatment on the disease progression~\cite{Linden2018}. Following the standard evaluation procedure, one would have to first randomly assign days to either the pre-lockdown or the lockdown condition and then estimate the impact of policies by calculating the mean difference. Obviously, random assignment of days to the condition is not feasible due to the inverse causality. Therefore, different approaches are implemented in the literature. 

The most common is based on the comparison of the measured air pollution over the lockdown period in 2020 to the same time interval in 2019. However, air pollution is known to depend on weather conditions, their recent history, season as well as policy updates that prohibit an accurate estimation of pollution reductions due to COVID-19 lockdown interventions. In this section we provide pollution reduction estimates that take weather-related parameters into account. In the next section, we introduce a way to refine these estimates using the measurements during the lockdown.

Local weather highly impacts the daily change of air pollution. In \figref{fig:weather-comparison}, we compare the weather conditions during the lockdown period in 2020 and during the same period in 2019. We observe that in Austria,  variation of temperature was higher in 2020 than in 2019. For Switzerland we see that the lockdown weather was warmer, dryer, and less windy than in 2019. Similar observations apply to Beijing and Wuhan. In addition, we notice a change in the wind direction, which is a significant pollution predictor in these regions due to a strong pollution transfer phenomenon~\cite{zhang2019multi}. The significant role of wind in the Beijing and Wuhan models is also reflected in \tabref{tab:model_selection_count} by a high number of the \preLD GAM models where WS, WX and WY were chosen as important explanatory variables by the model selection algorithm.

To estimate pollution reduction during the lockdown, we leverage the \preLD models trained on the pre-LD data as outlined in \secref{sec:longtermmodel}, and use these to predict air pollution concentrations over the lockdown period. We then compute the difference between the predicted and actually measured values to estimate the impact of the lockdown measures in each region.
Overall, the estimated pollution change over the LD period compared to the same time period in 2019 evaluates up to -29.4\% / +9.8\% and -22.6\% / +11.5\%, change in NO2 / PM10 in Eastern Switzerland and Lower Austria respectively;   -29.4\%, / -10.8\,\% and -52.8\,\% / -50.0\,\% in NO2 / PM2.5 in Beijing and Wuhan

\begin{figure*}[t]
    \centering
    \subfloat[\footnotesize{NO2, High traffic}]{
		\includegraphics[width=.49\textwidth]{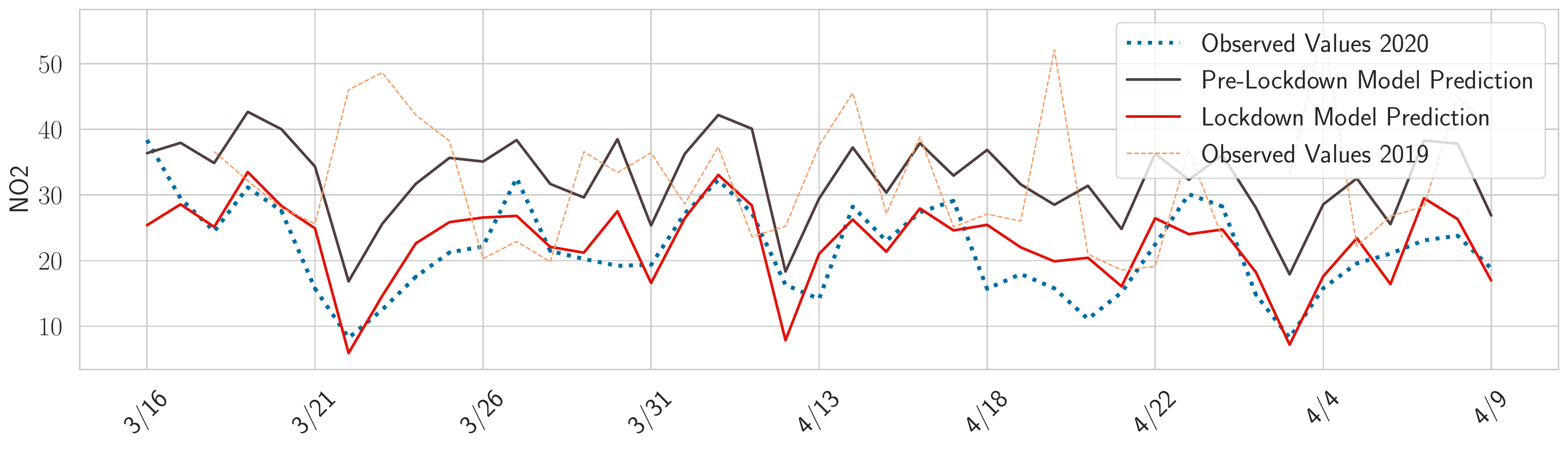}}
    \subfloat[\footnotesize{NO2, Low traffic}]{
		\includegraphics[width=.49\textwidth]{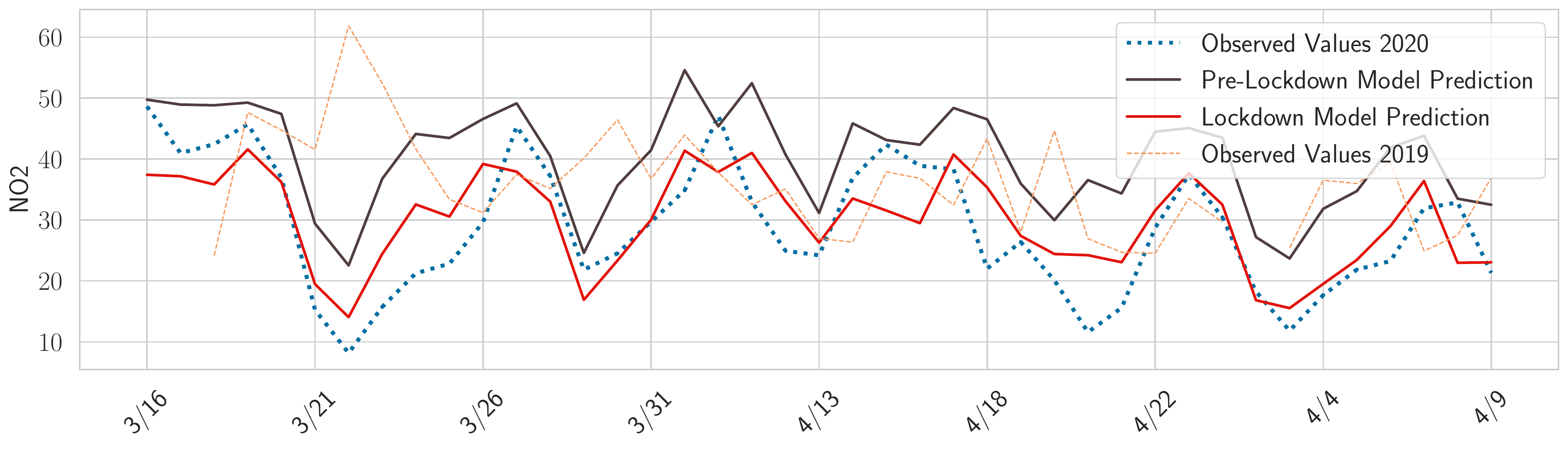}}
	\\
	\subfloat[\footnotesize{PM10, High traffic}]{
		\includegraphics[width=.49\textwidth]{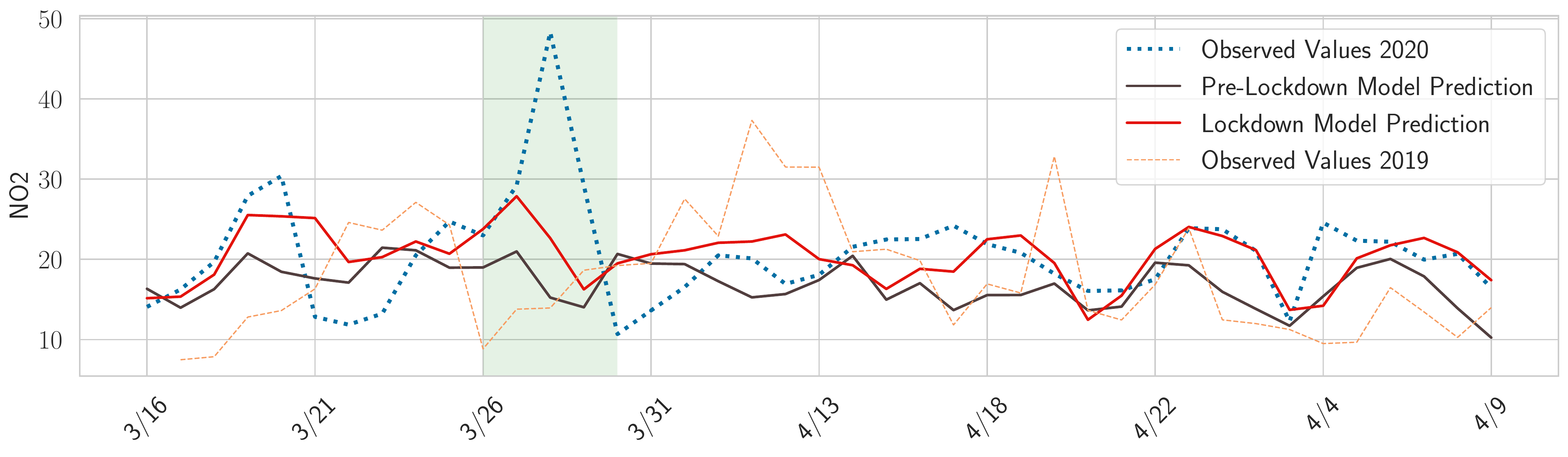}}
    \subfloat[\footnotesize{PM10, Low traffic}]{
		\includegraphics[width=.49\textwidth]{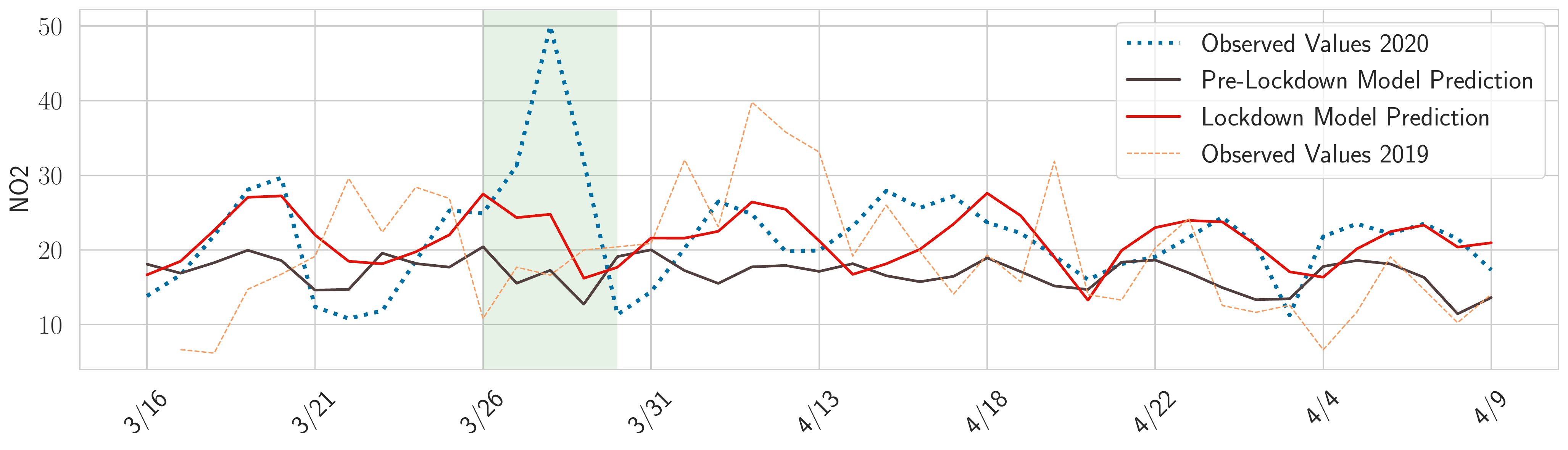}}
    \caption{Sample predictions for NO2 and PM10 for Eastern Switzerland. Green zones show the Sahara Dust Storm period~\cite{SRF_Sahara2}}.
    \label{fig:che-preLD-examples}
\end{figure*}

\begin{figure*}[!t]
    \centering
    \subfloat[\footnotesize{NO2, Road}]{
		\includegraphics[width=.49\textwidth]{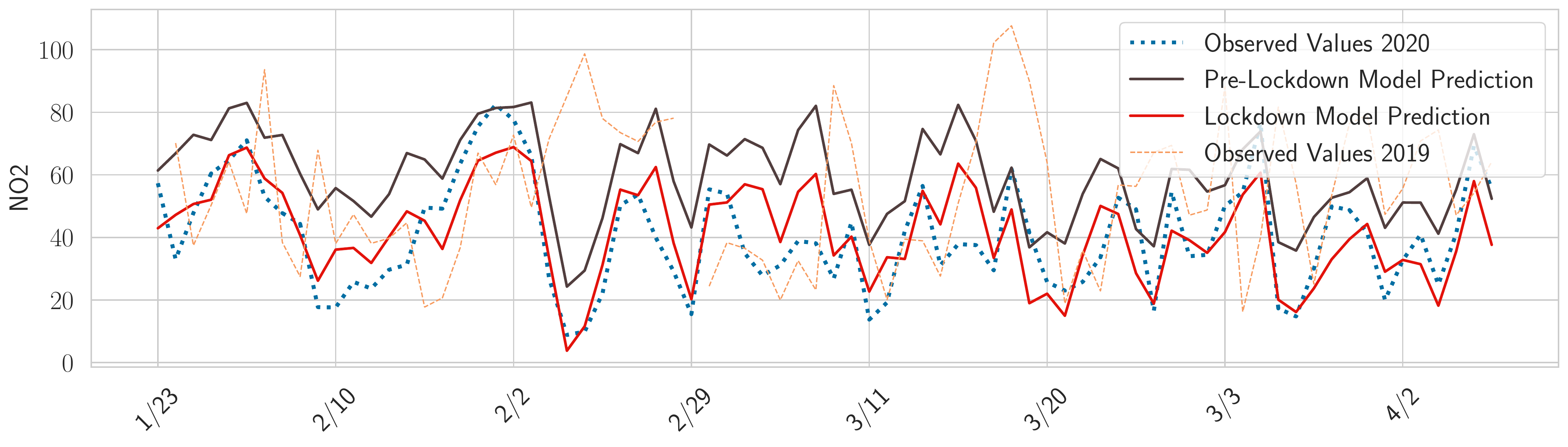}}
	\subfloat[\footnotesize{NO2, Rural}]{
		\includegraphics[width=.49\textwidth]{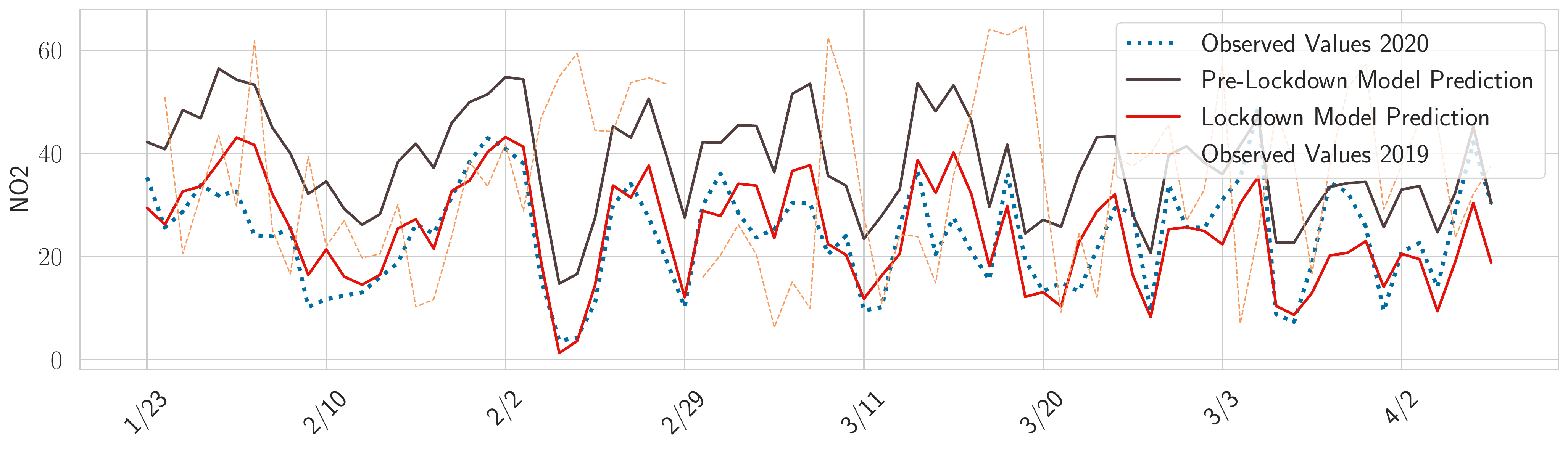}}
	\\
    \subfloat[\footnotesize{PM2.5, Road}]{
		\includegraphics[width=.49\textwidth]{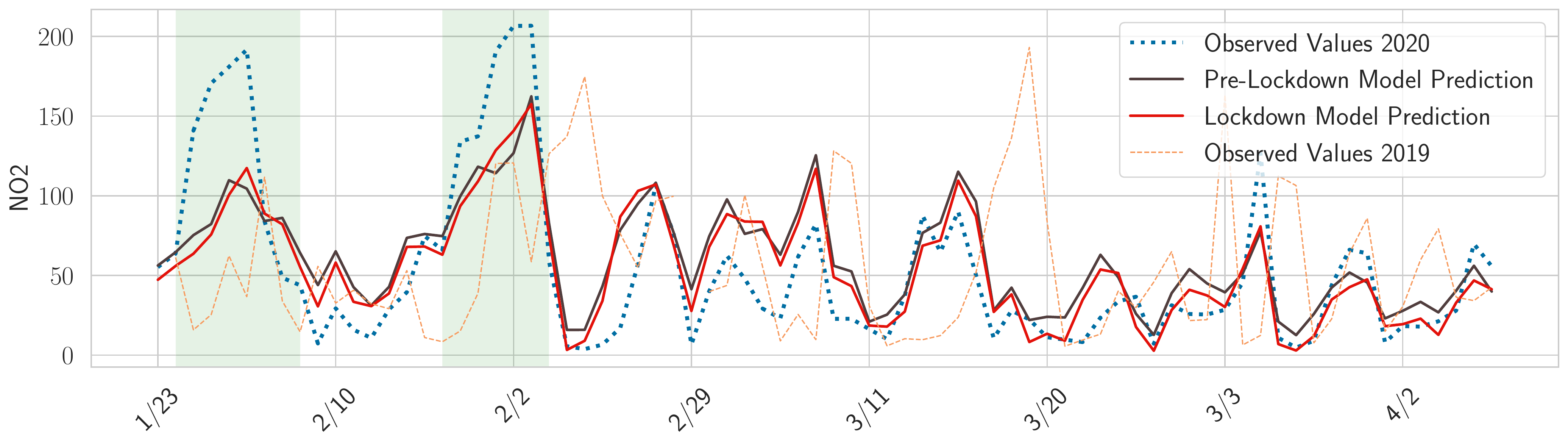}}	
    \subfloat[\footnotesize{PM2.5, Rural}]{
		\includegraphics[width=.49\textwidth]{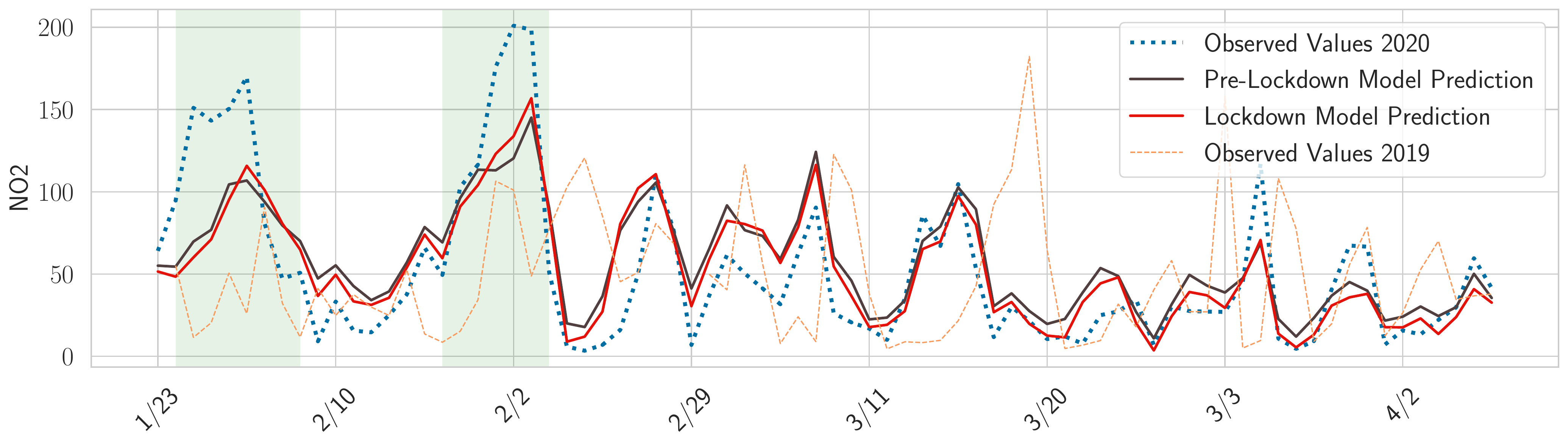}}	
    \caption{Sample predictions for NO2 and PM2.5 for Beijing, China. Green zones show the pollution transfer phenomenon, which caused unexpectedly high values in the measurements during the lockdown period in the beginning of 2020.}
    \label{fig:cn-preLD-examples}
\end{figure*}

 

\fakeparagraph{Eastern Switzerland}
Since NO2 is highly impacted by traffic, we put the obtained NO2 reduction estimates in the context of traffic reduction reported by Apple and Google based on the observed change in usage of their services.

The Apple Mobility Trends Report~\cite{Apple20} publishes aggregated estimates of the changed driving behavior of their users based on the navigation requests. The data suggests an average reduction of driving activity in Switzerland of 40.4\,\% compared to the baseline of Jan 13, 2020. As shown in \tabref{tab:che_apple},  this traffic drop translated into a reduction of NO2 between 78.4\,\% and 35.5\,\%  for Low Traffic and High Traffic stations respectively as predicted with our models, when using the same baseline. \figref{fig:che-preLD-examples} compares the output of the \preLD models to the observed values in 2019 and 2020 for these classes. For the station located off the road, we find an increase of 40.8\,\%. In this case, the  average NO2 is very low and any marginal change results in very high percentage variations. These likely can be attributed to weather phenomena not captured by our explanatory variables. A similar increase of 27.4\,\% was reported by \cite{Grange20} for another rural station in Switzerland.

Counter-intuitively, we found an increase of PM10 concentrations over the lockdown period despite reduced traffic intensity. However, an intense Sahara dust event occurred in Switzerland during the lockdown month starting on Mar 26, 2020~\cite{SRF_Sahara1, SRF_Sahara2}.  Sahara dust events are known to significantly increase PM concentrations in southern and middle Europe. Studies in Greece found Sahara dust contribution to PM2.5 of up to 82\,\% for severe events~\cite{Remoundaki13}. A similar investigation in Turkey~\cite{Kabatas14} indicates that Sahara dust is responsible for the increase of PM10 concentrations of up to 96\,\%. As can be seen in  \figref{fig:che-preLD-examples} the Sahara dust event in Switzerland resulted in a major increase in PM10 values which is not captured by our explanatory variables and explains the overall increase in PM10 despite reduced traffic.

\begin{table}[ht]
\small{
\subfloat[\preLD reduction estimates.]{
    \begin{tabular}{cccccc}
    \toprule
    \multicolumn{2}{c}{No Traffic}    & \multicolumn{2}{c}{Low Traffic}    & \multicolumn{2}{c}{High Traffic} \\

    NO2 & PM10 & NO2 & PM10 & NO2 & PM10 \\ 
    \% & \% & \% & \% & \% & \% \\ 
    \midrule
    +40.8    &+225  & -78.3   &+71.3    & -35.5  &+47.4 \\
    \bottomrule
    \end{tabular}
}
\hspace{.3cm}
\subfloat[Apple.]{
    \begin{tabular}{c}
    \toprule
    \multirow{2}{*}{Apple~\cite{Apple20}} \\
    \\
    \%  \\
    \midrule
    -40.38 \\
    \bottomrule
    \end{tabular}
}
\caption{Estimated pollution reduction in Eastern Switzerland compared to the traffic reduction reported by Apple~\cite{Apple20}.}
\label{tab:che_apple}
}
\end{table}

The Google Community Mobility Report~\cite{Google20} offers statistics on the estimated times spent in certain areas such as workplaces, groceries and parks. The statistic is derived from the user behavior using the Google Maps service. The median value of the period between Jan 3, 2020 and Feb 5, 2020 is used as a baseline for this calculation. The resulting estimates are listed in \tabref{tab:che_google} and indicate \eg a reduction of time spent in grocery stores of 13.1\,\% and a reduction of transit of 48.8\,\%. Using the same baseline, our model indicates a reduction of NO2 of 30.0,\% for Low Traffic stations and of 26.7\,\% for High Traffic. For the No Traffic station as well as for all classes for PM10, we again observe an increase due to the reasons explained above.

\begin{table*}[ht]
\small{
\subfloat[\preLD reduction estimates.]{
    \begin{tabular}{cccccc}
    \toprule
    \multicolumn{2}{c}{No Traffic}    & \multicolumn{2}{c}{Low Traffic}    & \multicolumn{2}{c}{High Traffic} \\
    NO2 & PM10 & NO2 & PM10 & NO2 & PM10 \\
    \% & \% & \% & \% & \% & \% \\
    \midrule
    +15.7 &+96.9 &-30.0 &+31.5 & -26.7 &+22.2 \\
    \bottomrule
    \end{tabular}
}
\hspace{1cm}
\subfloat[Google.]{
    \begin{tabular}{cccccc}
    \toprule
    \multicolumn{6}{c}{Google~\cite{Google20}} \\
    Recreation & Grocery & Parks & Transit & Workplaces & Residential \\
    \% & \% & \% & \% & \% & \% \\
    \midrule
    -72.2 & -13.1 &+11.0 & -48.8 & -43.8 & +20.5 \\
    \bottomrule
    \end{tabular}
}
\caption{Estimated pollution reduction in Eastern Switzerland compared to the mobility change as reported by Google~\cite{Google20}.}
\label{tab:che_google}
}
\end{table*}

We show example predictions for Low Traffic and High Traffic stations in Switzerland in \figref{fig:che-preLD-examples}. We observe that the measured NO2 and PM10 values differ significantly from the values in 2019 for the same time period. For the majority of days, measured NO2 concentrations lie below the \preLD model predictions. This is, however, not the case for PM10 for the reasons given above.

\fakeparagraph{Beijing and Wuhan}
We estimate an average NO2 reduction of 28.1\,\% and 52.8\,\% over the whole lockdown period for Beijing and Wuhan, respectively. A comparison of our weather-aware predictions to the published results~\cite{Bauwens20} obtained based on the analysis of the satellite images over the lockdown period (see TROPOMI~\cite{Bauwens20} and OMI~\cite{Bauwens20} is shown in \tabref{tab:cn_results_comparison}).  These estimates are for the same period as analysed in \cite{Bauwens20}, lasting from February 11th 2020 to March 3rd 2020. Our results for NO2 reductions  fall within the confidence intervals of the satellite imagery based predictions. 

\begin{table*}[ht]
\small{
\begin{tabular}{l|cccccccc|cc}
\toprule
 \multirow{3}{*}{Estimators}            & \multicolumn{2}{c}{Beijing--Urban}     & \multicolumn{2}{c}{Beijing--Suburban}  & \multicolumn{2}{c}{Beijing--Rural}     & \multicolumn{2}{c}{Beijing--Road}      & 
 \multicolumn{2}{c}{Wuhan}     \\

 & NO2 & PM2.5 & NO2 & PM2.5 & NO2 & PM2.5 & NO2 & PM2.5 & NO2 & PM2.5 \\
 & \% & \% & \% & \% & \% & \% & \% & \% & \% & \% \\ 
\midrule
\preLD                   & -34.7 &-26.2 &-27.6 &-43.5  &-38.6  &-34.6  &-34.8 & -23.8  &-62.4 &95.4 \\
TROPOMI~\cite{Bauwens20} &-25 $\pm$10 &- &-25 $\pm$ 10 &- & -25 $\pm$ 10 &- & -25 $\pm$ 10 &&-43 $\pm$ 14    &-    \\
OMI~\cite{Bauwens20}     &-33 $\pm$ 10 &- & -33 $\pm$ 10 &- & -33 $\pm$ 10 &- & -33 $\pm$ 10 &&-57 $\pm$14    &-  \\
\bottomrule
\end{tabular}
\caption{Estimated pollution reduction for Beijing and Wuhan compared to \cite{Bauwens20} as measured by satellite imagery.}
\label{tab:cn_results_comparison}
}
\end{table*}

Despite the estimated overall traffic reduction of 70\,\% for Beijing~\cite{Bauwens20}, the estimated PM2.5 concentrations over the lockdown decrease only slightly decrease by 10.8\% over the whole lockdown period. Similar results were also obtained for Beijing when using the Chemical Transport Model~\cite{zhang2012real}.
The explanation provided in the research literature attributes this to the impact of high ozone along with the considerably reduced NO2 that led to the  increased oxidation capacity in the ambient air~\cite{Ge17, Plautz18}. This resulted into more intense particle formation from the agricultural emissions, \eg ammonia (NH3), brought into the city with the wind. Our statistical model does not model pollutant interactions and thus does not capture these effects.
Interestingly, in Wuhan there is no evidence that a similar increase of oxidation capacity occurred during the lockdown. This observation is aligned with our predictions for Wuhan that show an estimated reduction of PM2.5 of an average 52.2\,\% over the lockdown period.

The average predictions for urban and rural stations in Beijing are exemplified in \figref{fig:cn-preLD-examples}. Measured NO2 values are significantly below \preLD predictions, whereas measured PM2.5 concentrations are very close to the predictions by the \preLD model. 
For the PM2.5 predictions, both the \LD and \preLD models fail to accurately predict two spikes in the beginning of the considered period, as shown in the green zones of \figref{fig:cn-preLD-examples}. The reason for this is the pollution transfer phenomenon: Air pollution is brought by the wind from the surrounding non-lockdown cities to Beijing~\cite{beijing_pm_reason}, and results in unexpectedly high measured air pollution concentrations in the city~\cite{zheng2015forecasting}.
Our proposed method could be easily adapted to acquire better prediction results by including prior knowledge about the pollution propagation patterns that are specific to Beijing~\cite{zhu2017pg}. This is, however, out of scope of this paper, yet would be interesting to model as part of future work. Apart from these two spikes highlighted in the plot, our proposed \LD model achieves reasonable prediction accuracy compared to the observed ground truth values in 2020.

\fakeparagraph{Lower Austria}
For Lower Austria, an overall decrease in NO2 of 22.6\% has been observed over the lockdown period. This closely matches the 23\% reduction that has been estimated for the city of Vienna in a recent study based on satellite data~\cite{Thieriot20}. As Lower Austria is surrounding Vienna, this is an indication that the developed model is in line with other research in this area. 
Regarding PM10, an increase of an average 11.5\% is observed during the lockdown period. Although, this might seem counter intuitive, meteorologists ascribe a large part of the variability of PM10 in Lower Austria to Sahara dust events and the still high number of wood fired ovens \cite{ORF20, Tech20}. As these variables are not included in the model, this can lead to deviations which are not explained by the model. Lockdowns further lead to an increase in the amount of time people spend at home which potentially led to heavier usage of wood fired ovens and thus raised PM10 levels.

Taking a closer look at the different classes of stations, shows that, as listed in Table \ref{tab:est_at} the NO2 reduction in rural areas amounts to only 8\% whereas the highest reduction of 33\%  has been observed at stations in an urban environment. With NO2 being highly affected by traffic volume, these results are in line with environmental theory.
For PM10 the opposite holds true, an increase of 17\% is observed in rural areas whereas no change of patterns (+0.4\%) occurred in cities. This is another indicator that the overall increase in PM10 can be attributed to wood fired ovens as these are more common in less urbanized parts of the country. 

\begin{table}
\small{
\begin{center}
\begin{tabular}{l|c|c|c|c}
\toprule
Class & NO2 & PM10 \\
\midrule
Residential             &   -24.2\%    &+6.8\% \\
Rural                   &  -8.2\%    &+16.7\% \\
Suburban Residential    & -17.3\%   &+18.5\% \\
Urban                   & -33\%  &+0.4\% \\
\bottomrule
\end{tabular}
\caption{Estimated pollution reduction in Lower Austria}
\label{tab:est_at}
\end{center}
}
\end{table}

\begin{figure*}[ht]
	\centering
	\subfloat[\footnotesize{Switzerland}]{
		\includegraphics[width=0.25\textwidth]{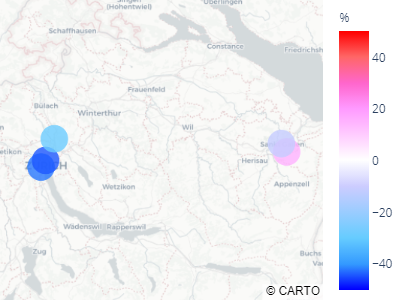}}
			\subfloat[\footnotesize{Austria}]{
		\includegraphics[width=0.25\textwidth]{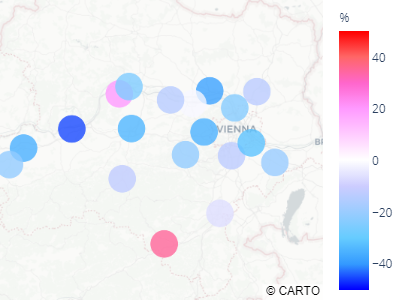}}
	\subfloat[\footnotesize{Beijing}]{
		\includegraphics[width=0.25\textwidth]{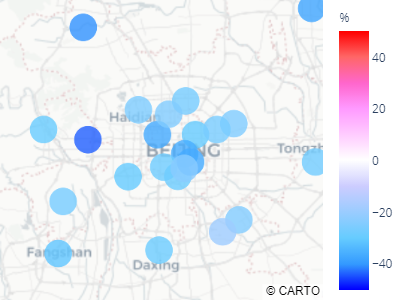}}
	\subfloat[\footnotesize{Wuhan}]{
		\includegraphics[width=0.25\textwidth]{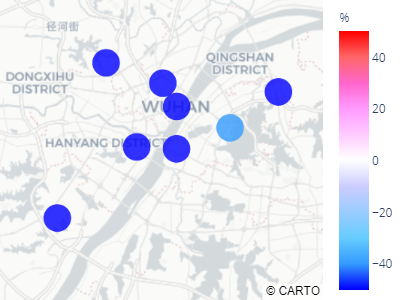}}
	\caption{Spatial distribution of pollution reduction for NO2 in Switzerland, Austria, Beijing and Wuhan}
	\label{fig:spatial-predictions}
\end{figure*}

We also provide the spatial distribution of the pollution reductions over the lockdown period across all stations in our data sets in \figref{fig:spatial-predictions}. When comparing major cities under analysis (Wuhan and Beijing in China; St.\,Gallen and Zurich in Eastern Switzerland), we conclude that NO2 and PM reductions were higher in cities with stricter enforced intervention measures. Given the influence of non-traffic related phenomena on PM2.5 and PM10 in Switzerland, Austria and Beijing, we consider only NO2 in further modelling. We note however that PM2.5 reduction in Wuhan was not affected by these events and confirms our expectations. We observe an average 95.4\,\% reduction of PM2.5 in Wuhan when comparing our \preLD model predictions to measured values (see \tabref{tab:cn_results_comparison}).

\begin{figure*}[ht]
	\centering
	\subfloat[\footnotesize{Switzerland}]{
		\includegraphics[width=0.25\textwidth]{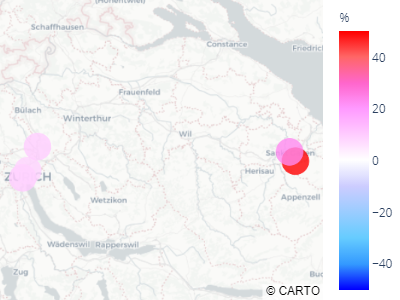}}
			\subfloat[\footnotesize{Austria}]{
		\includegraphics[width=0.25\textwidth]{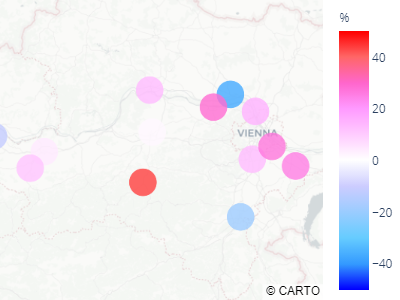}}
	\subfloat[\footnotesize{Beijing}]{
		\includegraphics[width=0.25\textwidth]{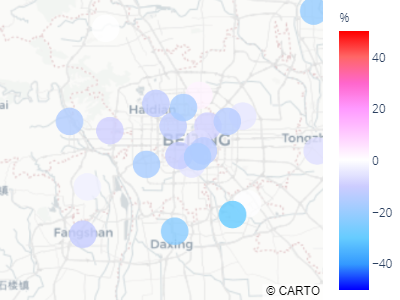}}
	\subfloat[\footnotesize{Wuhan}]{
		\includegraphics[width=0.25\textwidth]{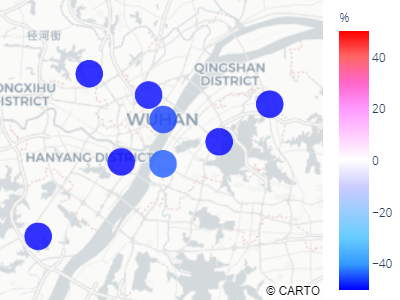}}
	\caption{Spatial distribution of pollution reduction for PM10 in Switzerland \& Austria and PM2.5 in Beijing and Wuhan}
	\label{fig:spatial-predictions2}
\end{figure*}




\section{Learning Lockdown Models}
\label{sec:transferability}

In the previous section we showed how to accurately estimate pollution reduction with a \preLD model by comparing the model predictions to the actually measured values. The lockdown period gives us the bottom-line by how much humans in different regions, given their cultural and political differences, can reduce their activities in a fear of getting infected by a virus. Having a bottom-line is useful when evaluating future policy changes, sector restructuring due to technological advances, process optimizations, \etc. In this section we describe the construction of the \LD models by transfer learning and show the value of both models in the analysis of the developments in the post-lockdown period.

Transfer learning is a popular technique to apply the knowledge gained by solving a particular talk to a related task~\cite{wei2016transfer}. Since the lockdown period was too short to fit GAM models for this time period, we apply transfer learning to \preLD models to derive models for the lockdown period. In this step we re-train the models on the scarce lockdown data to only fit the variables where we suspect the dependencies may have changed due to lockdown, \ie the day of the week. These variables serve as proxy for the traffic intensity. All weather dependencies in the \LD model are considered to be the same as in the \preLD model, which is confirmed by the domain experts. Therefore, the knowledge gained with regards to the influence of the weather and seasonality on air pollution can be transferred to the lockdown period. Thus, there is no need to train the entire model from scratch.

\begin{figure*}
    \centering
    \subfloat[\footnotesize{NO2, High traffic}]{
		\includegraphics[width=.46\linewidth]{graphics/che_no2/high_traffic_comparison.pdf}}
    \subfloat[\footnotesize{NO2, Low traffic}]{
		\includegraphics[width=.46\linewidth]{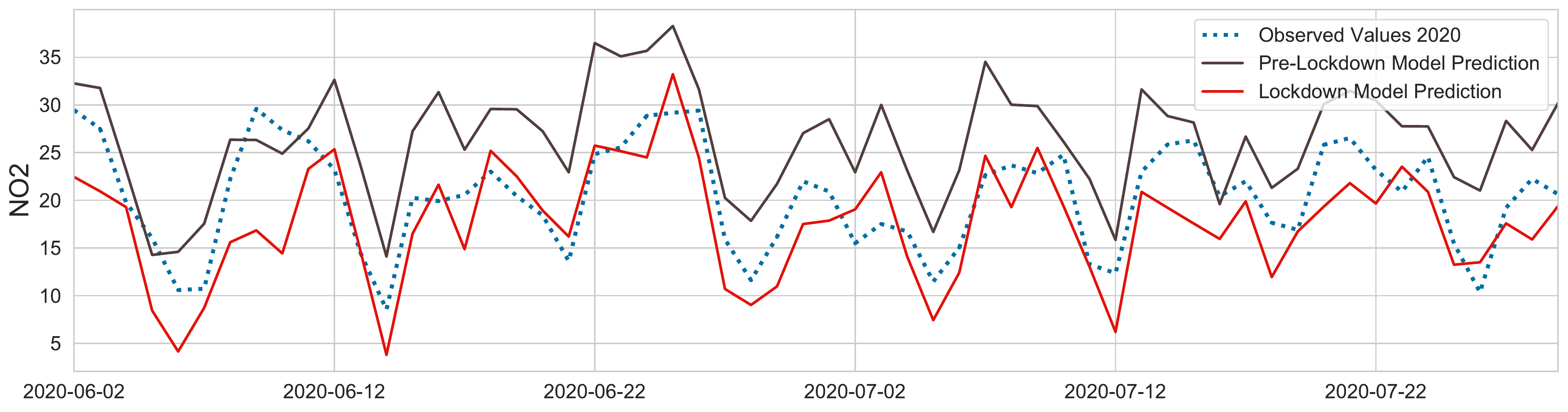}}
    \caption{Post-lockdown predictions, Eastern Switzerland.}
    \label{fig:ch-postlockdown}
\end{figure*}

\begin{figure*}
    \centering
    \subfloat[\footnotesize{NO2, Road}]{
		\includegraphics[width=.46\linewidth]{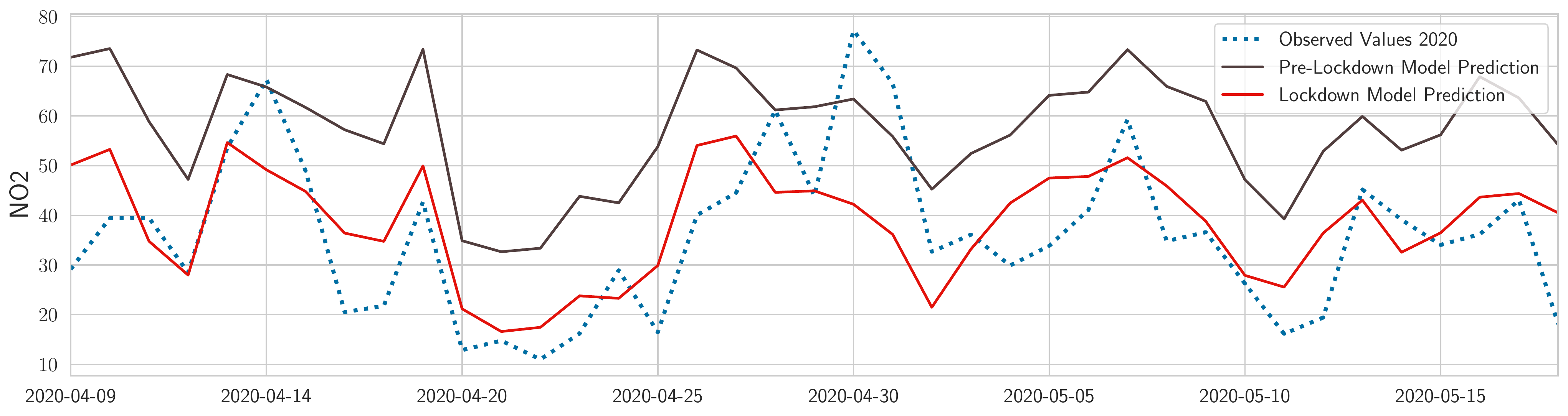}}
    \subfloat[\footnotesize{NO2, Rural}]{
		\includegraphics[width=.46\linewidth]{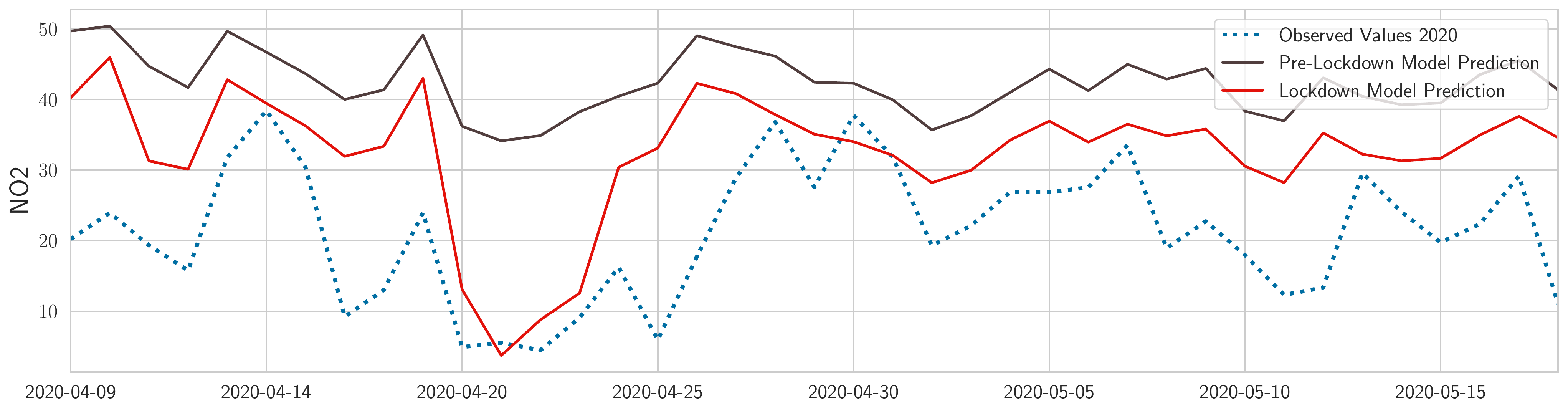}}
    \caption{Post-lockdown predictions, Beijing, China.}
    \label{fig:ch-postlockdown}
\end{figure*}

\subsection{Model Validation}


Due to the scarcity of measured data over the lockdown period, we validate the model by cross-validation using only 3 successive days as test data. The remaining data from the lockdown period is then used for training. This way, we get 14 estimates for the out of sample prediction RMSE. The summary of the average RMSE values for the lockdown model can be found in \tabref{tab:validation} and shows similar or better model performance for NO2 compared to the quality of the \preLD models.
For NO2 in Eastern Switzerland, we can clearly see that the \LD model closely matches the observed values. On average, the \LD models have a RMSE of 7.03 whereas the \preLD models have a RMSE of 7.16. This shows that the fine-tuned \LD model reflects well the dependency between air pollution and explanatory variables for the short lockdown time period. For Austria the model performance improvement is even more pronounced: The \preLD model reaches a RMSE of 4.5 while the \LD model achieves 3.7.  Similarly, for Beijing and Wuhan we obtain errors for \LD models of 13.08 and 12.26 for both cities respectively, compared to the RMSE of the \preLD model of 13.38 and 14.61. Model validation based on NO2 data shows that the \LD model does a good job in modelling the dependency change during lockdown in all areas. As next we show that the \LD model can indeed be useful to analyze the impact of traffic reduction on air pollution.

\begin{table}
\small{
\begin{center}
\begin{tabular}{l|c|c|c}
\toprule
\multirow{3}{*}{Region} & \multirow{3}{*}{Class}  & Hypothetical \\
 &  & reduction 2019 \\
 & & \% \\
\midrule
\multirow{3}{*}{Eastern Switzerland} & No Traffic &+0.7 \\
& Low Traffic   & -34.5 \\
& High Traffic & -29.4 \\
\hline
\multirow{3}{*}{Lower Austria} & Residential &-28.7 \\
& Rural  & -11.7 \\
& Suburban Residential & -20.6 \\
& Urban & -36.4 \\
\hline
\multirow{4}{*}{Beijing} & Urban  & -41.8 \\
& Rural         & -43.4 \\
& Suburban      & -36.9 \\
& Road &        -32.6 \\
\hline
Wuhan & Average & -60.4 \\
\bottomrule
\end{tabular}
\caption{Optimized mixture of \preLD and \LD model to explain observed NO2 values in the post-lockdown period.}
\label{tab:mixture}
\end{center}
}

\end{table}

\subsection{Evaluation of the Post-Lockdown Period}

We now use both models \preLD and \LD to investigate the optimal mixture thereof capable of explaining the observed pollution values after the lockdown period. By doing so we aim to estimate to what extent have human mobility and the inherent traffic gone back to normal. To run this analysis, we minimize the absolute sum of differences between the true observations and the mixture of the \preLD and \LD model predictions. 
\begin{equation}
\arg\min_\alpha \frac{1}{||T||} \sum_{t \in T} |\alpha \cdot m_t^{\text{\LD}} + (1-\alpha) \cdot m_t^{\text{\preLD}} - m_t|,
\label{eq:alpha}
    \end{equation}
\noindent where $m_t$ is the measured value at time $t \in T$, $T$ is a post-lockdown period of length $||T||$ under consideration, $m_t^{\text{\LD}}$ and $m_t^{\text{\preLD}}$ are the predictions obtained with \LD and \preLD models respectively. The dependent variable $\alpha$ shows the contribution of the \LD model when explaining the post-lockdown pollution measurements.
The results of this analysis for all areas are summarized below.

\begin{figure*}
    \centering
	\includegraphics[width=\linewidth]{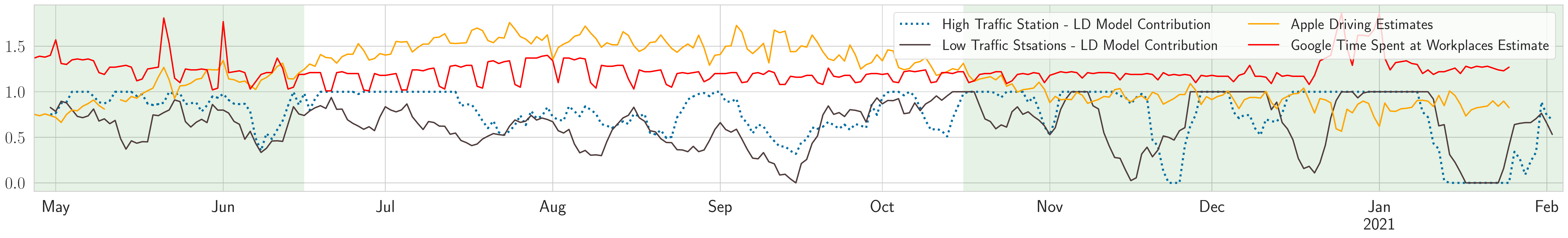}
    \caption{Contribution of the \LD model ($\alpha$ in \eqref{eq:alpha}), estimates of changes in driving from the Apple Mobility Report~\cite{Apple20} and estimated time spent at work places form the Google Community Mobility Report~\cite{Google20} compared to the report's respective baseline, for the time after the initial lockdown in Switzerland.}
    \label{fig:ch-postlockdown}
\end{figure*}

\begin{figure*}
    \centering
	\includegraphics[width=\linewidth]{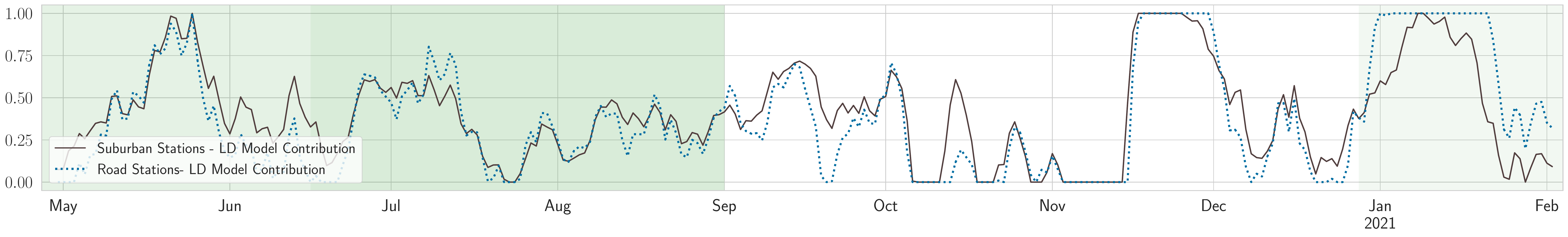}
    \caption{Contribution of the \LD model ($\alpha$ in \eqref{eq:alpha}) for the time after the first lockdown in Beijing.}
    \label{fig:chn-postlockdown}
\end{figure*}

\fakeparagraph{Eastern Switzerland}
For Eastern Switzerland we analyze the data after the initial lockdown, which covers the time period between May 2020 and Feb 2021. The results in \figref{fig:ch-postlockdown} show that the first months following the initial lockdown are close to the lockdown situation. Over summer the mobility gradually resumes back to $\sim$50\,\% normal again. However from Oct 2020 onwards, we can again observe a considerable traffic reduction due to mobility reducing measures put in place by the Swiss government in response to rising numbers of COVID cases~\cite{BAG1}. From Dec 22, 2020 until the end of our dataset (Feb 5, 2021), Switzerland has been put under a lockdown again~\cite{suedkurier21, BAG2}. However, our model estimates a low reduction this time (between 18\,\% and 24\,\%)for the traffic exposed stations. The reason for this is only a limited decrease of traffic density: According to the mobility estimates by Apple~\cite{Apple20}, driving in Switzerland reduced by on average 40\,\% throughout the initial lockdown in spring 2020, whereas during the second ongoing lockdown, only a 15\,\% reduction has been observed.

\fakeparagraph{Beijing}
%
In Beijing the second lockdown was imposed between June 15, 2020 and Sept 1, 2020. For this time period we estimate a reduction in NO2 of 11\,\% for rural, 15\,\% for urban, 7\,\% for suburban and 
23\,\%
for road stations. These comparatively small reductions are to be expected as the second lockdown hasn't been nearly as strict as the first one. As shown in \figref{fig:chn-postlockdown}, only in October, NO2 values came back to their typical pre-COVID-19 levels in Beijing. Since Dec 29, 2020 a light, partial lockdown has been in place in some areas of Beijing~\cite{Reuters21} which resulted in a sudden return to a level of 100\% lockdown model contribution, as highlighted in \figref{fig:chn-postlockdown}.

\subsection{Hypothetical Year-Long Reductions}

We can use the \LD model to predict the impact of the hypothetical traffic reduction policies on air quality. We demonstrate this using the year 2019 as an example and base the \LD model predictions using weather traces from the whole year. The estimated impact of a hypothetical year-long lockdown on NO2 reduction between Jan 1, 2019 and Dec 31, 2019 is compared to the measured values in 2019. The results are summarized in the last column of \tabref{tab:mixture}.

\fakeparagraph{Eastern Switzerland}
For the no traffic station in Eastern Switzerland we only find an increase of 0.7\%, which is plausible as this station is only affected by traffic to a very small degree. For Low Traffic and High Traffic we find an estimated reduction of 34.5\,\% and 29.4\,\%, respectively. These numbers show that also for a country with very low air pollution levels, traffic reduction can still have a considerable impact.

\fakeparagraph{Lower Austria}
For rural stations in Lower Austria reduction potential is only 11.7\% which is in line with our expectations as these stations are usually located far off big streets. The highest estimated reduction is found for urban areas, where our model suggests a 36.4\% decrease under lockdown conditions throughout 2019.

\fakeparagraph{Beijing and Wuhan}
For Beijing we observe an average estimated NO2 reduction between 43.4\,\% and 32.6\,\%. This shows a huge potential the traffic drop may have in cities suffering from toxically high air pollution such as Beijing. Also for Wuhan we estimate a reduction potential of 60.4\,\% if the lockdown measures would have been in place during the whole 2019. However, it must be taken into account that the lockdown in Wuhan was much stricter than in Beijing and thus overestimates the reduction potential if keeping mobility at a reasonably low level.

\section{Conclusion and Discussion}
\label{sec:conclusion}
 \enlargethispage{5pt} 
This paper proposes an approach to estimate the impact of the COVID-19 lockdown measures on local air quality as measured by ground measurement stations. Related works base their estimates of pollution reduction by comparing the values to the same period in 2019 or by the analysis of the satellite imagery. By contrast, our model learns a dependency between local air quality and weather augmented by the daytime specific dynamics impacted by the land-use and traffic. We build long-term prediction \preLD models using two years of historical data for the stations in Eastern Switzerland and China and compare the respective predictions to the measured values over the lockdown. Our analysis and findings match recent literature and show significant decrease in NO2 in all areas, yet a minor decrease or even an increase of PM10 and PM2.5 concentrations over the lockdown in Eastern Switzerland and Beijing, respectively. The reasons lead back to a major Sahara dust event spotted in Switzerland~\cite{SRF_Sahara1,SRF_Sahara2} over the lockdown period and an increased oxidation capacity in Beijing~\cite{beijing_pm_reason} that caused a higher than expected impact of the agricultural pollution from suburban areas brought to Beijing with the wind.

Due to a short lockdown duration it is impossible to learn a meaningful model for the lockdown period from scratch. To overcome the problem, we use transfer learning to re-train only a subset of explanatory variables on the scarce lockdown data. The resulting \LD models provide a bottom-line for the pollution reduction in various areas due to lockdown interventions of different strength. Both models match the quality of the state-of-art air pollution models described in the literature and were applied to the analysis of the post-lockdown period. We found that the society restored its mobility patterns in urban areas up to 58\,\% in Eastern Switzerland within two months following the lockdown. Mobility in Beijing was kept low up to 94\,\%, whereas the strictness of the lockdown in Wuhan was relaxed up to 65\,\%. These results should be interpreted in the context of varying severity of the lockdown measured in these areas. We also estimate a hypothetical whole-year pollution reduction for 2019 if the lockdown interventions would have been applied for a year. This would result in hypothetical 18.1\,\% and 54.2\,\% NO2 reduction in Eastern Switzerland and Beijing respectively. Thus, the proposed modelling framework can be used for the analysis of future policy changes in this context. 

 In this paper, we show how transfer learning can be used to update air pollution models should the weather relationships, policies or human activity change. Due to a known model structure and interpretability, it is easier to ensure validity and trustworthiness of such updates and support automated decision making and control. In the future, we plan to extend the \LD and \preLD models to spatial predictions by incorporating land-use data similar to \cite{Hasenfratz15}, and provide a visualization tool to show the effects of various policy measures and compare these to the lockdown baselines.



\vspace*{-0.2cm}

\bibliographystyle{ACM-Reference-Format}
\bibliography{literature}

\end{document}